\documentclass[10pt,journal,compsoc]{IEEEtran}
\usepackage[utf8]{inputenc}

\usepackage{float}
\usepackage{graphicx}
\usepackage{bm}
\usepackage{amsmath}
\usepackage{amsfonts}
\DeclareMathOperator*{\argmin}{argmin}
\usepackage{booktabs}
\usepackage{import}
\usepackage{gensymb}
\usepackage{color}
\usepackage{tikz}
\usepackage{pgfplots}

\usepackage{mathrsfs}
\usepackage{mathtools}

\usepackage{caption}

\definecolor{myblue1}{RGB}{0, 169, 255}
\definecolor{myblue2}{RGB}{0, 94, 255}
\definecolor{Fucsia}{RGB}{255, 0, 255}
\definecolor{myred}{RGB}{212, 0, 0}
\definecolor{mygreen}{RGB}{0, 212, 0}
\definecolor{airforceblue}{rgb}{0.36, 0.54, 0.66}
\newcommand{\bs}[1]{\boldsymbol{#1}}

\usepackage{booktabs} 

\usepackage{tikz}
\usepackage{stanli}
\usepackage{subcaption}

\begin{document}
\title{MORPH-DSLAM: Model Order Reduction for PHysics-based Deformable SLAM}

\author{A. Badias,
	I. Alfaro,                
        D. Gonzalez,
	F. Chinesta,
        E. Cueto%
\IEEEcompsocitemizethanks{\IEEEcompsocthanksitem A. Badias, I. Alfaro, D. Gonzalez, and E. Cueto are with the Instituto de Investigaci\'{o}n en Ingenier\'{i}a de Arag\'{o}n, Universidad de Zaragoza, Spain.\protect\\
E-mail: \{abadias, iciar, gonzal, ecueto\}@unizar.es
\IEEEcompsocthanksitem F. Chinesta is with the ESI Group Chair at the PIMM Lab, Arts et M\'{e}tiers Institute of Technology, Paris, France.\protect\\
E-mail: {francisco.chinesta@ensam.eu}}%
}

\IEEEtitleabstractindextext{%
\begin{abstract}
We propose a new methodology to estimate the 3D displacement field of deformable objects from video sequences using standard monocular cameras. We solve in real time the complete (possibly visco-)hyperelasticity problem to properly describe  the strain and stress fields that are consistent with the displacements captured by the images, constrained by real physics. We do not impose any ad-hoc prior or energy minimization in the external surface, since the real and complete mechanics problem is solved. This means that we can also estimate the internal state of the objects, even in occluded areas, just by observing the external surface and the knowledge of material properties and geometry. Solving this problem in real time using a realistic constitutive law, usually non-linear, is out of reach for current systems. To overcome this difficulty, we solve off-line a parametrized problem that considers each source of variability in the problem as a new parameter and, consequently, as a new dimension in the formulation. Model Order Reduction methods allow us to reduce the dimensionality of the problem, and therefore, its computational cost, while preserving the visualization of the solution in the high-dimensionality space. This allows an accurate estimation of the object deformations, improving also the robustness in the 3D points estimation.
\end{abstract}

}

\maketitle

\IEEEdisplaynontitleabstractindextext

\IEEEpeerreviewmaketitle

\IEEEraisesectionheading{\section{Introduction}\label{sec:introduction}}

\IEEEPARstart{H}{uman} visual sense is able to extract and process an enormous quantity of information, simply by observing a few images. A similar behavior is present in standard cameras, where an image can provide a high quantity of information. A strong development in computer vision has been carried out along last decades to extract relevant information from images. In this work, we want to focus in 3D reconstruction methods since our goal is to provide valuable information about the physical changes that happen in the surrounding space of an agent (a navigation robot or a user, employing augmented reality tools for the latter).

Visual augmented reality is based in the addition of graphical information to interact with reality, so it requires great efforts to fix virtual objects within the real world. This means that we need to know the map of the real world to locate the virtual objects. This process is known as Structure-from-Motion (SfM), and it consists in the estimation of rigid 3D structures from planar images using different camera view points.

Using the same tools, but with a different goal, we may want a robot to understand the surrounding spatial environment, with the ability of understanding the scene where it is moving. Sometimes not only geometric measurements from a sensor are needed, in some cases we should provide the agent with the ability of semantic perception and understanding of the physical changes that happen. We live in a dynamic world, where changes happen at any time, so robots need to have this comprehension ability to really understand the surrounding space.

\begin{figure}[h]
  \centering
  \centerline{\includegraphics[width=0.7\linewidth]{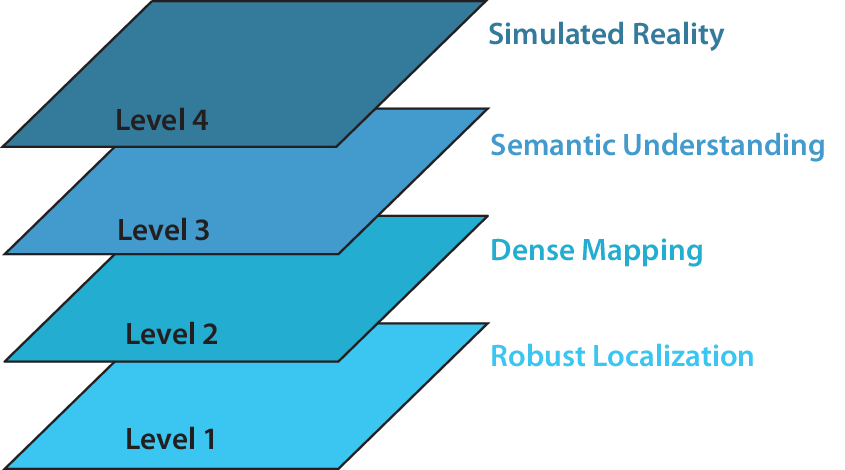}}
  \caption[Levels in Spatial AI.]{Levels in the Spatial AI approach. Image inspired in the CVPR presentation of A. Davison \cite{Davison2019CVPR}.}
  \label{fig:C1:SpatialAI}
\end{figure}

Some authors suggest the use of the term \emph{Spatial AI} \cite{Davison:2018aa}. It consists of a 3-layer hierarchical system for the spatial understanding of the environment \cite{Davison2019CVPR}, namely \emph{Robust Localization}, \emph{Dense Mapping} and \emph{Semantic Understanding}. In our work we suggest to add a fourth layer to understand the physical changes that occur around the agent (see Fig.~\ref{fig:C1:SpatialAI}). The physical understanding may be gathered within the \emph{semantic understanding}, {  but we consider that both are very relevant features and must be considered separately.}

The first level of the Spatial AI approach is related to the ability of the agent to locate itself in space and navigate through it in a robust manner. The great development in the last decades has brought new algorithms of \emph{Simultaneous Localization and Mapping} (SLAM) performing localization tasks very efficiently and robustly \cite{Mur-Artal:2015aa}. Level 2 refers to the dense mapping that reconstructs the environment with a high level of detail \cite{Newcombe:2011ab}. Level 3 requires the system to identify and recognize the objects around it and to be able to establish semantic relationships \cite{Chen:2017aa}. And finally, at level 4, we place the term defined as \emph{Simulated Reality} (SR) \cite{Stoica:2017aa}, where the system needs to know the physics of the surrounding environment in order to have a deeper understanding of the world. SR is based on continuously simulating the physical phenomena of the environment with which the agent interacts. Learning this kind of concepts is a long and complex process, since we must teach a machine how to interpret the physics.

In the field of computer graphics, recent investigations have focused in the development of simplified physics (see, to name but a few \cite{li2018learning,pouring}) in order to overcome obvious computational limitations. However, these simplified approaches provide results with no known accuracy bounds. Dealing with deformable solids, there are some physics engines to approximate the dynamics, but they hardly comply with basic conservation laws. We consider there is no better approach than really solving the (in)elasticity equations, so we suggest to solve the high-fidelity equations.  However, standard solvers are not able to work in real time, and even less on portable devices. Since we use visual cameras, the video frequencies are usually fixed in 30 or 60 frames per second, so this is why we suggest the use of Model Order Reduction (MOR) techniques to work in this range of frequencies. MOR methods are based on the compression of the data to work in a reduced space, but at the same time they must recover the original high-fidelity data in a fast way, fulfilling visualization frequencies, with a minimal loss in accuracy. Moreover, this loss is limited by well established error bounds for the vast majority of MOR techniques. The power of these techniques allows us to simulate realistic behaviors that describe and adapt perfectly to the changes that occur in the environment.

In this paper we propose the use of reduced-order numerical methods to estimate the behavior of real deformable solids. Certainly it is usually thought that solving the equations of solid mechanics can be a heavy process, not suitable for video frequencies. But in recent years some tools have been developed to reduce the complexity of these mechanical models. They allow to approximate the true physics of solids at real-time rates.

Including this introduction section, we have divided this paper into 8 parts. Section 2 includes other relevant works in this field from a computer vision perspective. Section 3 explains our method in a few words. Section 4 shows the mechanical formulation of the problem. Section 5 shows the dimensionality reduction problem and the formulation of some MOR methods. Section 6 explains the inverse problem assimilating the data with a deformable implementation of ORB-SLAM2. Section 7 covers the experimentation phase. And finally, section 8 contains the conclusions of the work.

\section{Related Work}

The standard approach to compute rigid SfM is based in bundle adjustment (BA). This technique tries to minimize the reprojection error between observed objects and projected objects (observed in more than one image), allowing the estimation of the 3D points and camera locations \cite{Triggs:1999aa}. SfM method is normally a post-processing technique giving, in general words, more accurate results as much information is extracted from images. But some applications, such as automatic navigation, need an on-line mapping in real time. This means that the agent needs to compute SfM method in real time to be able to navigate in an unknown world, receiving the name of Simultaneous Localization And Mapping (SLAM), involving tasks of estimating the 3D environment structure at the same time the camera is located in the map.

Many works have been developed in this area, allowing to use different sensors (monocular \cite{Davison:2007aa}, stereo \cite{Paz:2008aa}, RGB-D \cite{Gutierrez-Gomez:2016aa} or laser technology \cite{Cole:2006aa}, among others). We can make a division in visual SLAM between direct \cite{Engel:2014aa} and feature-based \cite{Mur-Artal:2015aa}. Direct approaches use directly the image pixels minimizing the photometric error, while feature-based systems extract features from the images and use them as keypoints to apply BA techniques. Another classification can be {  made} between filtered \cite{Smith:1990aa} or keyframe-based \cite{Klein:2007aa} techniques, resulting the first idea in a filtering technique based in the propagation of probability functions and the second idea in a sparse implementation requiring graph optimizations. The work carried out for the last decades has allowed these techniques to move from only research to the development stage, appearing some companies that offer consumer tools based in SLAM techniques.

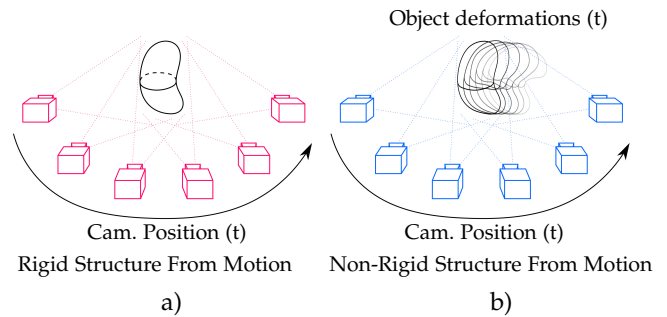
\begin{figure}[!h]
\centering
\def\svgwidth{0.45\textwidth}
\begingroup%
  \makeatletter%
  \providecommand\color[2][]{%
    \errmessage{(Inkscape) Color is used for the text in Inkscape, but the package 'color.sty' is not loaded}%
    \renewcommand\color[2][]{}%
  }%
  \providecommand\transparent[1]{%
    \errmessage{(Inkscape) Transparency is used (non-zero) for the text in Inkscape, but the package 'transparent.sty' is not loaded}%
    \renewcommand\transparent[1]{}%
  }%
  \providecommand\rotatebox[2]{#2}%
  \newcommand*\fsize{\dimexpr\f@size pt\relax}%
  \newcommand*\lineheight[1]{\fontsize{\fsize}{#1\fsize}\selectfont}%
  \ifx\svgwidth\undefined%
    \setlength{\unitlength}{1024.04221903bp}%
    \ifx\svgscale\undefined%
      \relax%
    \else%
      \setlength{\unitlength}{\unitlength * \real{\svgscale}}%
    \fi%
  \else%
    \setlength{\unitlength}{\svgwidth}%
  \fi%
  \global\let\svgwidth\undefined%
  \global\let\svgscale\undefined%
  \makeatother%
  \begin{picture}(1,0.47829574)%
    \lineheight{1}%
    \setlength\tabcolsep{0pt}%
    \put(0,0){\includegraphics[width=\unitlength]{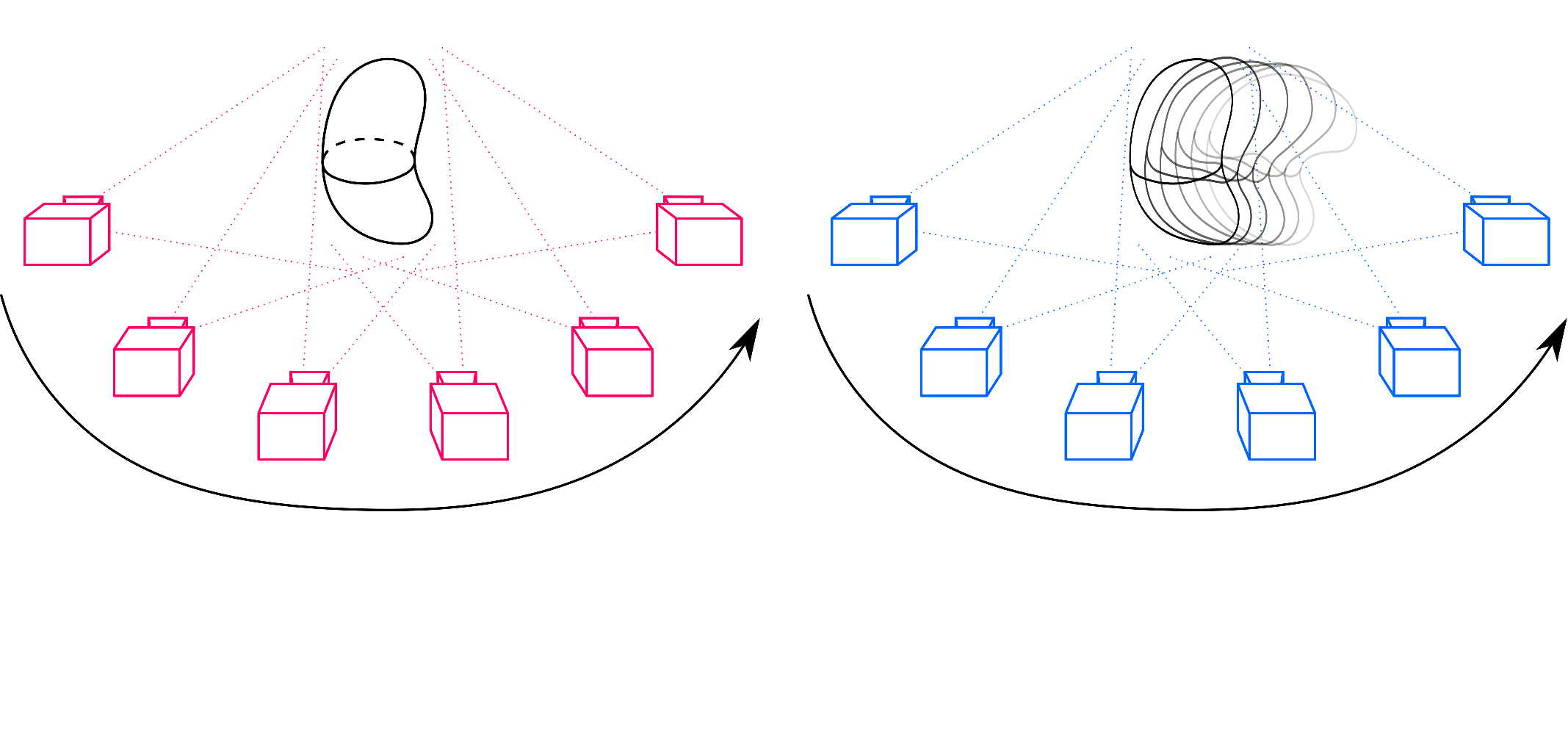}}%
    \put(0.23032575,0.06814018){\color[rgb]{0,0,0}\makebox(0,0)[t]{\lineheight{0}\smash{\begin{tabular}[t]{c}\footnotesize{Rigid Structure From Motion}\end{tabular}}}}%
    \put(0.77542047,0.06814018){\color[rgb]{0,0,0}\makebox(0,0)[t]{\lineheight{0}\smash{\begin{tabular}[t]{c}\footnotesize{Non-Rigid Structure From Motion}\end{tabular}}}}%
    \put(0.24962293,0.12030611){\color[rgb]{0,0,0}\makebox(0,0)[t]{\lineheight{0}\smash{\begin{tabular}[t]{c}\footnotesize{Cam. Position (t)}\end{tabular}}}}%
    \put(0.76461707,0.12030611){\color[rgb]{0,0,0}\makebox(0,0)[t]{\lineheight{0}\smash{\begin{tabular}[t]{c}\footnotesize{Cam. Position (t)}\end{tabular}}}}%
    \put(0.7878583,0.46716682){\color[rgb]{0,0,0}\makebox(0,0)[t]{\lineheight{0}\smash{\begin{tabular}[t]{c}\footnotesize{Object deformations (t)}\end{tabular}}}}%
    \put(0.25742035,0.00411971){\color[rgb]{0,0,0}\makebox(0,0)[t]{\lineheight{0}\smash{\begin{tabular}[t]{c}a)\end{tabular}}}}%
    \put(0.79255452,0.00411971){\color[rgb]{0,0,0}\makebox(0,0)[t]{\lineheight{0}\smash{\begin{tabular}[t]{c}b)\end{tabular}}}}%
  \end{picture}%
\endgroup%

\caption{ a) Rigid SfM allowing a reconstruction of the world from different views. b) Non-rigid SfM implies that both the camera and the scene may take different positions and shapes (time-dependent).}
\label{FigFig1}
\end{figure}

Another classification, with special interest for us, can be made regarding the position of the objects in the scene. Standard implementations of SfM (and SLAM) assume a rigid world, but as we said, sometimes this assumption cannot be made. The term non-rigid structure from motion (NRSfM) appeared to take into account this kind of movements, and it assumes the camera can move inside the world but also scene objects can move (or deform), as we can see in the right side of Fig.~\ref{FigFig1}. This is an ill-posed problem as we are not able to build the 3D objects from 2D points triangulation. In fact, a different configuration (non-rigid object deformation) can be observed in each image, {  making the direct estimation of the geometries difficult.}

During the last years, many works regarding NRSfM have been developed trying to solve this problem. Some of them used factorization approaches to obtain low-rank representations of the objects from image streams \cite{Bregler:2000aa}, where the 3D shape in each frame is a linear combination of a set of basis shapes (previously applied to rigid shapes by \cite{Tomasi:1992aa}). Also the non-rigid movements have been computed as a union of subspaces to model complex motions with clustering local subspaces \cite{Zhu:2014aa}. Some other works computed the trajectory space by a linear combination of basis trajectories instead of basis shapes using \emph{generic} bases \cite{Akhter:2009aa}. Non-linear dimensionality reduction methods have also been used to model the 3D shape as a non-linear combination of basis shapes, using kernel functions \cite{Gotardo:2011aa}, allowing a reduced number of bases if the problem is non-linear and the kernel is well defined. Also bayesian implementations have been used \cite{Agudo_etal_cviu2018}.

Priors were introduced to narrow down the search for an optimal solution \cite{Torresani:2008aa}, and some other works allow a prior less estimation of the structure from motion assuming compressible 3d objects as a block sparse dictionary learning problem \cite{Kong:2016aa}, and also applying dense variational reconstructions requiring GPU acceleration \cite{Garg:2013aa}. 

Other works reconstruct the shape of a deformable object using templates previously created \cite{Bartoli:2015aa}, obtaining accurate results, but requiring 2D parameterizations as they work only with the external surface, also with curved geometries \cite{Gallardo2020}. It means some energy constraints need to be imposed to assure boundary continuity and isometric deformations. Other works estimate the surface normals to help in the deformation tracking \cite{Lamarca2019}. And also simple physics-based implementations have tried to solve the motion of the external surface using physical priors and Kalman filter formulations \cite{Agudo:2016aa}.

There is also another approach that involves the use of the classical theory of finite element methods to estimate and track deformable objects with some advantages like robustness and real forces estimation in large deformations, applied to 1D objects \cite{Ilic:2007aa} but also to more complex surfaces \cite{Metaxas:1991aa, McInerney:1993aa}.

{  To the best of our knowledge, there is no alternative approach, however, in which the physics of the system under consideration is solved rigorously. This is due, undoubtedly, to the high computational cost of solving a high-fidelity model by employing state-of-the-art techniques such as finite elements or finite differences. This motivates our choice of reduced-order models, whose accuracy as a function of the complexity of the model is nowadays well-known and can be controlled by the analyst.}

\section{Overview of the Proposed Method}

{ To overcome the just mentioned difficulties, we propose to track deformable solids from video sequences by solving the (possibly non-linear) elasticity problem.} The goal is to create an agent able to understand the physical deformations of the objects to finally show some information to a user (e.g. stress or strain fields, or deformations of internal and hidden areas). We make use of reduced-order and parametric models, with which we solve the inverse problem to estimate the kinematic (displacements and strain fields) and dynamic (forces, stress) states, along with the rest of parameter values. Equivalently, the problem can be viewed as an example of data assimilation, in which we obtain parameter values by measuring displacements in the object external surface. Our approach is based on a two-stage problem. First, in an off-line stage we precompute a reduced-order, multiparametric approximation of the displacement field of the solids. This can be viewed as the estimation of a response surface. Second, an on-line stage is carried out to optimize the parameter values that better approximate the measured displacements in the video sequence under real time constraints. 

We do not impose any artificial spatio-temporal restriction or prior, nor {\em ad hoc} energy laws, since we are solving the actual physics in a finite element framework, guaranteeing energy conservation---dissipative behaviors do not impose any additional conceptual difficulty to this method---and elastic (or hyperelastic) constitutive laws. Of course, initial and boundary conditions need to be applied in the computation step, but they can also be considered as parameters in the parameterized problem. The proposed method can be applied both to articulated and continuous solids, beating any other existing method in terms of frequency rates (since it only requires an \emph{evaluation} of the parametric solution, rather than a true simulation), accuracy and robustness (noisy observations are minimized without the need of Bayesian implementations). Our method has no dependency on the acquisition system used (monocular, stereo or RGB-D systems), but we think, however, that one of the biggest advantages appears with the use of monocular sensors, {  since the introduction of parametric models helps in the minimization process, converging on an admissible solution.}

Strictly speaking, we can say that our procedure is not a structure-from-motion method, since we are not building a 3D structure from the motion data. The proposed method is related to the {\em Shape from Template} family of methods \cite{Bartoli:2015aa}, in which we register an off-line (reduced-order) parametric model to the camera observations. 

\section{Problem Formulation}

\begin{figure*}[!h]
\centering
\def\svgwidth{0.8\linewidth}
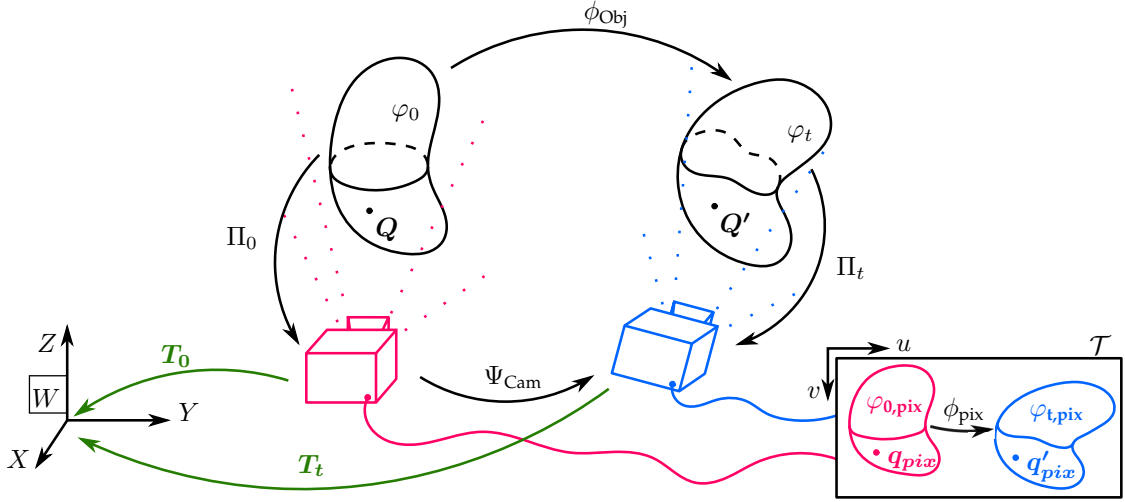
\caption{Kinematic modeling of the proposed formulation.}
\label{FigGeomModeling}
\end{figure*}

Very often the problem of NRSfM has been formulated respect to the motion of the visible portion of the deforming solid (the domain for which some information is captured by the camera). However, it is well-known that the motion of the visible part of the solid is heavily influenced by the whole geometry, including internal details, as described by the Navier equations of solid mechanics. Therefore, in this paper we formulate the problem as a complete, three-dimensional description of the solid \cite{Bonet:2008aa}, thus avoiding other NRSfM formulations focused only in the visible part of the solid whose motion was modeled by shell finite elements \cite{Agudo:2016aa}, \cite{Bartoli:2015aa}. We are interested in some problems like, for example, augmented reality for mechanical design or surgery guidance, some problems that allow us to assume that a full three-dimensional model is available off-line.

\subsection{Kinematic description of deforming solids}

Neglecting inertial effects and assuming, for simplicity, that there is no change in temperature, we consider a continuous solid $\Omega$, where we identify a point $\bs{Q}$, with coordinates $\bs{X}=(X,Y,Z)$ respect to \emph{World} (\emph{W} in Fig.~\ref{FigGeomModeling}). These coordinates are defined in the {\em undeformed configuration} $\varphi_{0} : \Omega \to \mathbb{R}^3$. The object undergoes a deformation so that point $\bs{Q}$ moves to the new position $\bs{Q'}$ with coordinates $\bs{X}'=(X',Y',Z')$, respect to \emph{W}, in the deformed configuration $\varphi_{t} : \Omega \to \mathbb{R}^3$. The function that explains the movement of the object is $\phi_{\text{Obj}}: \Omega \times \mathcal{I} \to \mathbb{R}^{3}$, {  where $\mathcal{I} \in \mathbb{R}$ is} the time interval where the movement is produced. Therefore, making use of a Lagrangian description of the amount of movement (material, always following the same infinitesimal particle), we can express the deformed configuration of point $\bs{Q}$ as
\begin{equation*}
\bs{X'} = \phi_{\text{Obj}} (\bs{X},t),
\end{equation*}
{  where $t$ is} the instant of time elapsed between both states. The deformation gradient tensor $\bs{F}$ (a two-point tensor) is the fundamental quantity that explains the deformation produced between two neighboring particles between the initial and the deformed state (assuming continuity in the mapping function). It is defined, with respect to the initial configuration, as
\begin{equation*}
\bs{F} = \frac{ \partial \bs{X'} }{\partial \bs{X}} = \frac{ \partial \phi_{\text{Obj}}(\bs{X},t) }{\partial \bs{X}} = \bs{\nabla} \phi_{\text{Obj}}.
\end{equation*}

It is convenient to define a strain measurement independent of the type of movement (rigid-solid translation or rotation, relative deformations or the combination of all of them). For independence with respect to to rotations, we use the \emph{right Cauchy-Green deformation tensor}, defined as
\begin{equation*}
\bs{C} = \bs{F^{T}} \bs{F}.
\end{equation*}

And finally, using the \emph{Green-Lagrange strain tensor} (material frame of reference) we can measure the deformations independently of rigid-body motions
\begin{equation}
\bs{E} = \frac{1}{2} (\bs{C} - \bs{I} ) = \frac{1}{2} \Big( (\bs{\nabla U})^T + \bs{\nabla U} + (\bs{\nabla U})^T  \bs{\nabla U} \Big),
\label{Eq:E}
\end{equation}
with  $\bs U(\bs X)= \bs{X'}-\bs X$ the material {  displacement.}

\subsection{Camera-centered kinematic description}

Let us assume that we obtain an image of the boundary of an opaque solid at the undeformed configuration, $\partial \Omega_{0}$. We consider a standard perspective projection camera \cite{Hartley:2003aa} (after intrinsic calibration and lens distortion correction) applying a transformation $\Pi_{0} : \partial \Omega_{0} \to \mathcal{T}$, {  where $\mathcal{T} \in \mathbb{R}^{2}$ represents} the image space. This camera operator $\Pi_{0}$ maps a point $\bs{Q}$ from \emph{World} coordinates to $\bs{q}_{\text{pix}}$ in \emph{pixel} coordinates $\bs x_{\text{pix}}=(u,v)$ making use of the camera intrinsic and extrinsic parameters. The same process occurs with projection $\Pi_{t} : \partial \Omega_{t} \to \mathcal{T}$, where points in the deformed configuration $\varphi_{t}$ are projected to the image $\varphi_{t,\text{pix}} \subset \mathbb{R}^{2}$.

\subsection{Solving the unknown kinematics}

Using a monocular implementation, the unknown mapping functions in our problem for any time instant $t \in \mathcal{I}$ are
\begin{itemize}
\item The displacements of the current configuration in a material description (i.e., with respect to the reference configuration) $\phi_{\text{Obj}}(\bs{X},t)$.
\item The camera projection $\Pi_{t}$, but only the extrinsic component $\bs{T}_{t}$, since it is the part that can change with the camera movement (and also the redundant function $\Psi_{\text{Cam}} : \mathbb{R}^3 \to \mathbb{R}^3 $, that maps the movement of the camera with respect to its initial position).
\end{itemize}

\begin{figure}[!h]
\centering
\def\svgwidth{0.18\textwidth}
\begingroup%
  \makeatletter%
  \providecommand\color[2][]{%
    \errmessage{(Inkscape) Color is used for the text in Inkscape, but the package 'color.sty' is not loaded}%
    \renewcommand\color[2][]{}%
  }%
  \providecommand\transparent[1]{%
    \errmessage{(Inkscape) Transparency is used (non-zero) for the text in Inkscape, but the package 'transparent.sty' is not loaded}%
    \renewcommand\transparent[1]{}%
  }%
  \providecommand\rotatebox[2]{#2}%
  \newcommand*\fsize{\dimexpr\f@size pt\relax}%
  \newcommand*\lineheight[1]{\fontsize{\fsize}{#1\fsize}\selectfont}%
  \ifx\svgwidth\undefined%
    \setlength{\unitlength}{229.71428098bp}%
    \ifx\svgscale\undefined%
      \relax%
    \else%
      \setlength{\unitlength}{\unitlength * \real{\svgscale}}%
    \fi%
  \else%
    \setlength{\unitlength}{\svgwidth}%
  \fi%
  \global\let\svgwidth\undefined%
  \global\let\svgscale\undefined%
  \makeatother%
  \begin{picture}(1,1.49774366)%
    \lineheight{1}%
    \setlength\tabcolsep{0pt}%
    \put(0,0){\includegraphics[width=\unitlength]{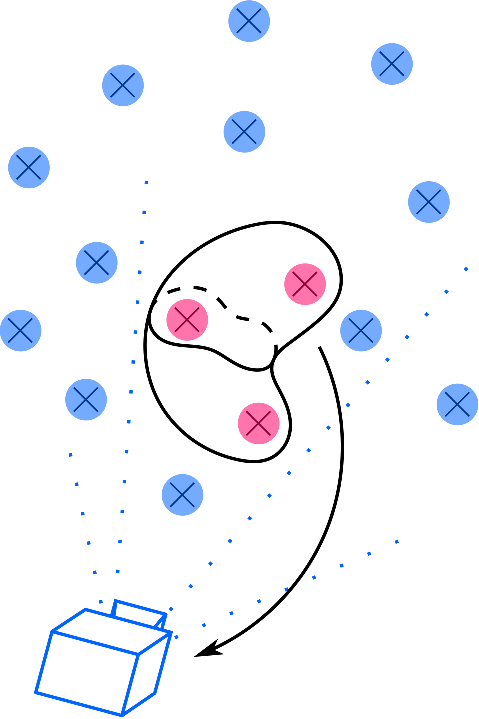}}%
    \put(0.81250692,0.4586666){\color[rgb]{0,0,0}\makebox(0,0)[t]{\lineheight{0}\smash{\begin{tabular}[t]{c}$\Pi_{i}$\end{tabular}}}}%
    \put(0.67320343,1.09249747){\color[rgb]{0,0,0}\makebox(0,0)[t]{\lineheight{0}\smash{\begin{tabular}[t]{c}$\varphi_{i}$\end{tabular}}}}%
  \end{picture}%
\endgroup%

\caption{In general, we consider applications where an important part of the scene is static (blue points), while some regions (red points) are non-rigid or deformable.}
\label{FigInliersOutliers}
\end{figure}

The estimation of the mapping $\Pi_{t}$ for each frame capturing rigid scenes is known, as we said, as the Structure-from-Motion (SfM) problem and it involves the estimation of the extrinsic parameters $\bs{T}_{t}$ (and the three-dimensional position of the points of the object, $\bs Q_{t}$), by minimizing a reprojection error in $L_{2}$-norm, commonly. If we also add the estimation of $\phi_{\text{Obj}}$, where objects can deform at any time instant, the problem becomes more difficult (ill-posed). It is possible that a different deformed configuration appears at every instant, with different camera positions. In this paper we assume that the camera captures both static and deformed points. Rigid or static points are related to the general scene (static objects, blue points in Fig.~\ref{FigInliersOutliers}) and are used to estimate the pose of the camera, and non-static points are used to determine the deformations of the non-rigid objects in the scene (red points in Fig.~\ref{FigInliersOutliers}). { We assume all deformable objects have fixed boundary conditions, so at least one point of the boundary is fixed in the scene. However, we could also assume that the objects, in addition to be deformed, move or rotate around the scene, in which case it would be necessary to estimate the relative position of the camera with respect to the object solving a similar but double problem (PnP problem \cite{Lepetit2009} and displacements).}

The position and orientation (pose) of the camera $\bs{T}_{t}$ is estimated by using fixed points (image features in our case) along with consensus techniques such as RANSAC \cite{RANSAC} to identify the subset of fixed points in the scene. To estimate the camera pose, the reprojection error $d(\bs{x}_{\text{pix}},\bs{\hat{x}}_{\text{pix}})$ is minimized, with $d$ the euclidean distance between the pixel coordinates of a point observed in an image and the pixel coordinates of the point projected on the image (from the 3D point $\bs{x}$ and the camera projection $\Pi_{t}$, parameters that are also optimized). We use ORB-SLAM2 \cite{Mur-Artal:2017aa} for the estimation of the camera pose and the static mapping.

The next task is to estimate the object deformations from its projection in the images. We use the red points in Fig.~\ref{FigInliersOutliers}, or in other words, the \emph{outliers}. RANSAC techniques applied to SLAM usually neglect those points that do not remain fixed in the scene. {  But our true  interest lies} in the information associated with the deformation of the objects. It is located precisely there, in the outliers. Therefore, these points provide the most valuable information about the displacements suffered by the objects, which we have to identify after carrying out a registration of the objects.

\subsection{Equilibrium in the deformable solid}
In previous sections we included a brief summary of the kinematic analysis of the deformable solid, but to fully resolve the mechanical problem it is necessary to take into account the stress analysis to guarantee equilibrium. If we split the solid by an arbitrary plane of unit normal $\bs n$, a force per unit area appears in any point of the newly created surface,
\begin{equation*}
\bs{t}(\bs{n}) = \bs{\sigma} \bs{n}; \hspace{15pt} \bs{\sigma} = \sum_{i,j=1}^{3} \sigma_{ij} \bs{e}_{i} \otimes \bs{e}_{j}
\end{equation*}
where $\bs{t}$ is the stress vector, $\bs{n}$ is the vector normal to the plane and $\bs{\sigma}$ is the \emph{Cauchy stress tensor}. 

The stress magnitude that is {\em conjugate} to the gradient of deformation tensor $\bs{F}$ (their product gives an energy measure) is the so-called \emph{first Piola-Kirchhoff stress tensor} $\bs{P}$, defined as
\begin{equation*}
\bs{P} = J \bs{\sigma} \bs{F}^{-T},
\end{equation*}
{  where $J = \det(\bs{F})$ is} the Jacobian of the transformation $\phi$. The stress measure conjugate to tensor $\bs{E}$ is the \emph{second Piola-Kirchhoff stress tensor} $\bs{S}$ defined as
\begin{equation*}
\bs{S} = J \bs{F}^{-1} \bs{\sigma} \bs{F}^{-T},
\end{equation*}
so we can write equivalently
\begin{equation*}
\bs{P} = \bs{F} \bs{S}.
\end{equation*}
In order to guarantee equilibrium in the solid, we must satisfy the following equilibrium equation, in the absence of inertia terms (assuming a quasi-static process):
\begin{equation}
\bs{\nabla} \cdot \bs{P} + \bs{B} = \bs{0} \text{ in } \Omega_t,
\label{Eq:equilibrium}
\end{equation}
{  where $\bs{B}$ are} the applied body forces. These equations need to be supplemented with appropriate boundary conditions,
\begin{align*}
\left\{ \begin{array}{rll}
\bs{\sigma} \hspace{2pt} \bs{n} & = \bs{\bar{t}} & \text{ on } \Gamma_{t}, \\
\bs{u} & = \bs{\bar{u}} & \text{ on } \Gamma_{u} \\
\end{array} \right.
\end{align*}
where $\bs{\bar{t}}$ are the traction forces acting along the boundary region $\Gamma_t$ in terms of the unit normal to the boundary, $\bs{n}$, and finally $\bs{\bar{u}}$ are the prescribed displacements defined in $\Gamma_u$. The \emph{weak form} of Eq.~(\ref{Eq:equilibrium}) is created to relax continuity and derivability restrictions on the functions that approximate the final solution, and is obtained after multiplying by an arbitrary function (admissible variation) and integrating by parts,
\begin{equation}
\int_{\Omega} \bs{S} : \delta \bs{E} \hspace{3pt} d\Omega = \int_{\Gamma_{t}}	\bs{\bar{t}} \hspace{3pt} \delta u \hspace{3pt} d\Gamma.
\label{Eq:WeakForm}
\end{equation}

\subsection{Constitutive equations}

The last ingredient to describe the whole mechanical process is to define its constitutive nature. {  In this work we consider hyperelastic material laws instead of linear elasticity, as linear materials may introduce artificial increments in volume when large deformations occur.} Hyperelastic materials are those whose stress-strain relationship derives from a strain energy density function $W$. For these materials, the second Piola-Kirchhoff tensor $\bs{S}$ can be obtained from the strain energy density function $W$ as
\begin{equation}
\bs{S} = \frac{\partial W(\bs E)}{\partial \bs E}.
\label{Eq:DerivEnergy}
\end{equation}
Adding Eq.~(\ref{Eq:DerivEnergy}) into Eq.~(\ref{Eq:WeakForm}), the problem to solve by using finite element methods \cite{Fish:2007aa} is
\begin{equation}
\int_{\Omega} \delta \bs{E}:\bs{\mathsf C}: \bs{E} \hspace{3pt} d\Omega = \int_{\Gamma_{t}}	\bs{\bar{t}} \hspace{3pt} \delta u \hspace{3pt} d\Gamma,
\label{Eq:WeakForm2}
\end{equation}
where $\bs{\mathsf C}$ is the fourth-order constitutive tensor. Section~\ref{Sec:Experims} shows a practical example of hyperelastic materials.

\section{Model Order Reduction (MOR) methods}

One of the most important aspects of our work is that we are assuming that the movement $\phi_{\text{Obj}}$ is bounded and can be parameterized and projected onto a low-dimensional manifold. This means that the objects we simulate do not have the infinite number of degrees of freedom typical of continuum mechanics. Due to the laws of continuum mechanics, strong correlations exist between the displacement suffered by each material point. Actually, we can express the solution as a function of a much smaller number of degrees of freedom, what we call dimensions in our reduced space. 

\subsection{Dimensionality Reduction}
\label{Sec:DimReduction}

The computational complexity of the continuum mechanics problem prevents its solution under real-time constraints. Note the non-linear character imposed by the strain measures introduced in Eq.~(\ref{Eq:E}), and also the common non-linear terms of hyperelastic laws. To guarantee that we comply with video frequencies, a reduction of the complexity of the problem is mandatory. In addition, the parameter dependency (material properties, geometry parameters or boundary conditions, among others) increases the complexity of the problem.

Let us assume that the governing equation depends on a vector of parameters $\bs{\mu} \in \mathcal{P}$, where $\mathcal{P}$ is the set of all possible values of the parameters, and a compact subset of $\mathbb{R}^{\tt n_{\text{param}}}$. The manifold or solution set is $\mathcal{M}$, where all solutions $\bs{U}(\bs{\mu})$ remain,
\begin{align*}
\mathcal{M} = \varphi (\mathcal{P}) = \{\bs{U}(\bs{\mu}) \in V : \bs{\mu} \in \mathcal{P} \subset \mathbb{R}^{\tt n_{\text{param}}} \},
\end{align*}
where $\varphi$ is again the solution map and $V$ is a suitable Hilbert space. The solution map $\varphi$ is defined as
\begin{align*}
\varphi : \mathcal{P} \rightarrow V \text{,} \hspace{20pt} \bs{\mu} \mapsto \bs{U}(\bs{\mu}).
\end{align*}

It is not possible to work with the original solution set due to the finite memory of computers, so we have to make a simplification by choosing a suitable discretization technique to obtain the high-fidelity solution set $\mathcal{M}_{h}$, also known as the discrete manifold, usually obtained by finite element techniques. We assume that $\mathcal{M}_{h}$ is so close to $\mathcal{M}$ that very small differences appear at the discretization points. The discretized solution of the equation is $\bs{U}_{h}(\bs{\mu})$ and it belongs to a finite-dimensional subspace $V_{h}$ of dimension $N_{h}$,
\begin{align*}
\mathcal{M}_{h} = \varphi_{h} (\mathcal{P}) = \{ \bs{U}_{h}(\bs{\mu}) \in V_{h} : \bs{\mu} \in \mathcal{P} \} \subset V_{h}.
\end{align*}
After applying the approximation to $\bs U(\bs \mu)$, we obtain also the discrete solution map
\begin{align*}
\varphi_{h} : \mathcal{P} \rightarrow V_{h} \text{,} \hspace{20pt} \bs{\mu} \mapsto \bs{U}_{h}(\bs{\mu})
\end{align*}

There are many MOR methods to reduce the dimensionality of the problem with the goal of obtaining a reduced basis with an optimal number of components. The reduced solution thus belongs to a low-dimensional subspace $V_{r} \subset V_{h}$ of dimension $N_{r} \ll N_{h}$ where the construction of the basis of $V_{r}$ is generated from a set of $r$ functions. The particular form of constructing these functions can be separated between \emph{a posteriori} and \emph{a priori} methods. The classical methods are \emph{a posteriori}, where some data is coming from an unkown source and statistical tools like Principal Component Analysis \cite{Karhunen:1947aa,Loeve1963} are applied to project the data in a reduced and more efficient space. The Proper Orthogonal Decomposition (POD) method \cite{Berkooz:1993aa} is a common example. But, if a multiparametric problem has to be solved, the curse of dimensionality appears, so \emph{a priori} methods can be applied to solve the mechanical problem directly in the reduced space, without evaluating the whole set of solutions in the high-dimensionality space.

To better understand the procedure, we detail next the type of model order reduction methods commonly applied to continuum mechanics and how they can be of great help in our work.

\subsection{MOR in Computational Mechanics}
\label{MORinCM}

We are looking for a suitable approximation of a space, time and parameter-dependent field $\bs U(\bs X,\bs \mu, t)$. Under the term {\em model order reduction} we encompass a family of methods that look for an approximation of the type
\begin{equation}\label{sep2}
\bs U(\bs X,t,\bs \mu) \approx \sum_{i=1}^{{\tt n}_{\text{MOR}}} \bs F_i(\bs X) \circ \bs G_i (t)\circ \bs H_i(\bs \mu),
\end{equation}
where $\bs F_i(\bs X)$, $\bs G_i(t)$ and $\bs H_i(\bs \mu)$ represent the so-called {\em modes} of the approximation, i.e., ${\tt n}_{\text{MOR}}$ vector-valued functions approximated in a finite element sense that best approximate the unknown field $\bs U$. The symbol ``$\circ$'' represents the Hadamard or Schur (component-wise) product of vectors.

The simplest way to determine an optimal approximation of the type given by Eq. (\ref{sep2}), sometimes referred to as {\em affine} or {\em separate}, is by applying Principal Component Analysis of a set of snapshots of the evolution of the system, called Proper Orthogonal Decomposition in computational mechanics \cite{Wilcox2}. Once the modes have been identified, by injecting the approximation given by Eq. (\ref{sep2}) into the weak form of the differential equation governing the problem, the reduced system can be solved.

Other methods like Reduced Basis (RB) employ directly a set of snapshots of the system, whose parameter value is chosen under well-defined error criterion \cite{Quarteroni:2015aa}. 

On the other hand, Proper Generalized decompositions (PGD) \cite{ARCH11} \cite{Cueto:2016aa} do not employ any snapshot of the system. This approach presents several advantages. First of all, it is not necessary to solve the problem for all of the parameter combinations in the high-fidelity space. The problem is solved in the separated space, Eq. (\ref{sep2}), and it can be computed off-line and stored advantageously in memory in the form of a collection of vectors. These vectors include only the nodal values, at mesh vertices, of the functions $\bs F_i$, $\bs G_i$ and $\bs H_i$. At any other point of the model, the value of the field $\bs U$ is obtained by finite element interpolation. Secondly, this separated representation presents important advantages for the computation of inverse problems. The sensitivities (parameter influence in the solution) can be computed straightfoward thanks to this precise form of the approximation, very important to boost the optimization step. A detailed formulation is given in Section \ref{Sec:DefORBSLAM}.

\subsection{Proper Generalized Decompositions (PGD)}
As we said, PGD method does not need any observation of the system to construct the low-rank approximation to the unknown field. The method is based in two stages: an off-line step where the computation of the low-rank approximation given by Eq.~(\ref{sep2}) is performed; and an on-line phase where the parametric solution is evaluated under real-time constraints. To better explain the basics of PGD, we consider an introductory example. As a toy model problem we consider linear elasticity: the case of a cantilever beam in which the position of the applied load is considered as a parameter of the system.

\subsubsection{How PGD works: linear elasticity}
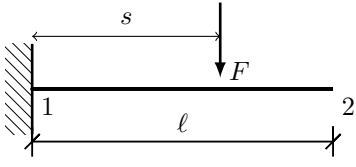
\begin{figure}[h!]
\centering
\begin{tikzpicture}
\point{a}{0}{0};
\point{b}{4}{0};
\point{c}{2.5}{0};
\point{d}{3}{1};
\point{e}{1.25}{0.75};
\beam {2}{a}{b}[1];
\support{3}{a}[270];
\notation {1}{b}{$2$}[below right];
\notation {1}{a}{$1$}[below right];
\dimensioning {1}{a}{b}{ -.7}[$\ell$];
\load{1}{c}[90];
\notation {1}{c}{$F$}[above right];
\notation {1}{e}{$s$}[above];
\draw[<->] (0,0.7)--(2.5,0.7);
\end{tikzpicture}
\caption{Cantilever beam problem. A moving load $F$ is parameterized by its position coordinate $s$.}\label{Fig:ExampleLE}
\end{figure}

Let us assume a vertical and constant force $F$, applied over the upper surface of a cantilever beam. The unknown field of the problem is the displacement $\bs{U}(\bs X, s)$, which depends on the particular material position $\bs X$ and on the position of the load, $s$ (see Fig.~\ref{Fig:ExampleLE}). The load can be applied at any point of the upper part $s \in \bar{\Gamma} \subset \partial \Omega$, with $\bar{\Gamma}$ the portion of the boundary $\Gamma_{t}$ with non-vanishing Neumann conditions.

The strong form of the governing equation (linear elasticity) under infinitesimal strain theory \cite{Fish:2007aa} is
\begin{align}
\bs U(\bs X,s) = \left\{ \begin{array}{rll}
\bs{\nabla \cdot \sigma} + \bs{B} & = \bs 0  & \text{ in }\Omega,  \\
\bs{\sigma} \hspace{2pt} \bs{n} & = \bs{\bar{t}} & \text{ on }\Gamma_{t}, \\
\bs{U} & = \bs{\bar{U}} & \text{ on }\Gamma_{U}, \\
\end{array} \right.
\label{Eq:StrFormLinElas}
\end{align}
where $\bs{\sigma}$ is the stress tensor, $\bs{B}$ is the body force per unit volume (assumed vanishing in this example, for the sake of simplicity), $\bs{\bar{t}}(s)$ are the traction forces applied to the solid and $\bs{\bar{U}}$ are the prescribed displacements (clamped support at point $1$). The constitutive equations for elastic materials, also known as \emph{Hooke's law}, represents the material behavior and relates stress and strain as
\begin{align*}
\bs{\sigma} = \bs{\mathsf{C}} {:} \bs{\varepsilon},
\end{align*}
with $\bs{\mathsf{C}}$ the fourth-order constitutive tensor and $\bs{\varepsilon}$ the (small) strain tensor. Assuming a hyperelastic framework (of which linear elasticity is a particular example), this constitutive tensor can be obtained by differentiating the strain energy density function $W$, Eq.~(\ref{Eq:DerivEnergy}), that in this case takes the form
\begin{align*}
W(\bs{\varepsilon}) = \frac{1}{2} E \bs{\varepsilon}^2,
\end{align*}
where parameter $E$ is the Young's modulus. Finally, the strain-displacement (kinematic) equations are
\begin{align*}
\bs{\varepsilon} = \frac{1}{2} [ \bs{\nabla U} + (\bs{\nabla U})^{T} ].
\end{align*}

Keeping in mind that $\bs U$ depends on $\bs{X} = (X,Y)$ and $s$, the weak form of the governing equation under the PGD formalism can be described as
\begin{align}
\int_{\bar{\Gamma}} \int_{\Omega} \bs{\nabla_{s}} \bs{U^{*}} \bs{:} \bs{\sigma} \hspace{2pt} d\Omega \hspace{2pt} d\bar{\Gamma} = \int_{\bar{\Gamma}} \int_{\Gamma_{t}} \bs{U^{*}} \hspace{2pt} \bs{t} \hspace{2pt} d\Gamma \hspace{2pt} d\bar{\Gamma}
\label{EqCantileverBeam}
\end{align}
where $\bs{U^{*}}  \in H^{1}_{0}$ is an arbitrary test function in the appropriate Sobolev space of functions vanishing along the Dirichlet boundary of the solid and $\bs{\nabla_{s}} = \frac{1}{2} [ \bs{\nabla} + (\bs{\nabla})^{T} ]$ is the symmetric gradient operator. The solution $\bs U(\bs{X},s)$ is assumed to be expressed in separate form,
\begin{align}\label{aa}
\bs U(\bs{X},s) \approx \sum_{i=1}^{{\tt n}_{\text{MOR}}} \bs F_{i}(\bs{X})\circ \bs G_{i} (s),
\end{align}
where ${{\tt n}_{\text{MOR}}}$ is the number of sums (rank of the tensor decomposition of the solution), $\bs F_{i}(\bs{X})$ are the space modes in separate variables and $\bs G_{i}(s)$ are the modes referred to the position of the load. An alternating direction scheme is used to obtain the ${{\tt n}_{\text{MOR}}}$ functions $\bs F_{i}(\bs{X})$ and $\bs G_{i}(s)$ so that, at iteration $p$ we search the $p+1$ functions $\bs R(\bs{X})$ and $\bs S(s)$ that enrich the solution
\begin{align*}
\bs U_{p+1}(\bs{X},s) = \bs U_{p}(\bs{X},s) + \bs R(\bs{X})\circ \bs S(s).
\end{align*}

\begin{figure}[H]
\footnotesize{
\centering
\begin{subfigure}{\linewidth}
\centering
\def\svgwidth{0.825\textwidth}
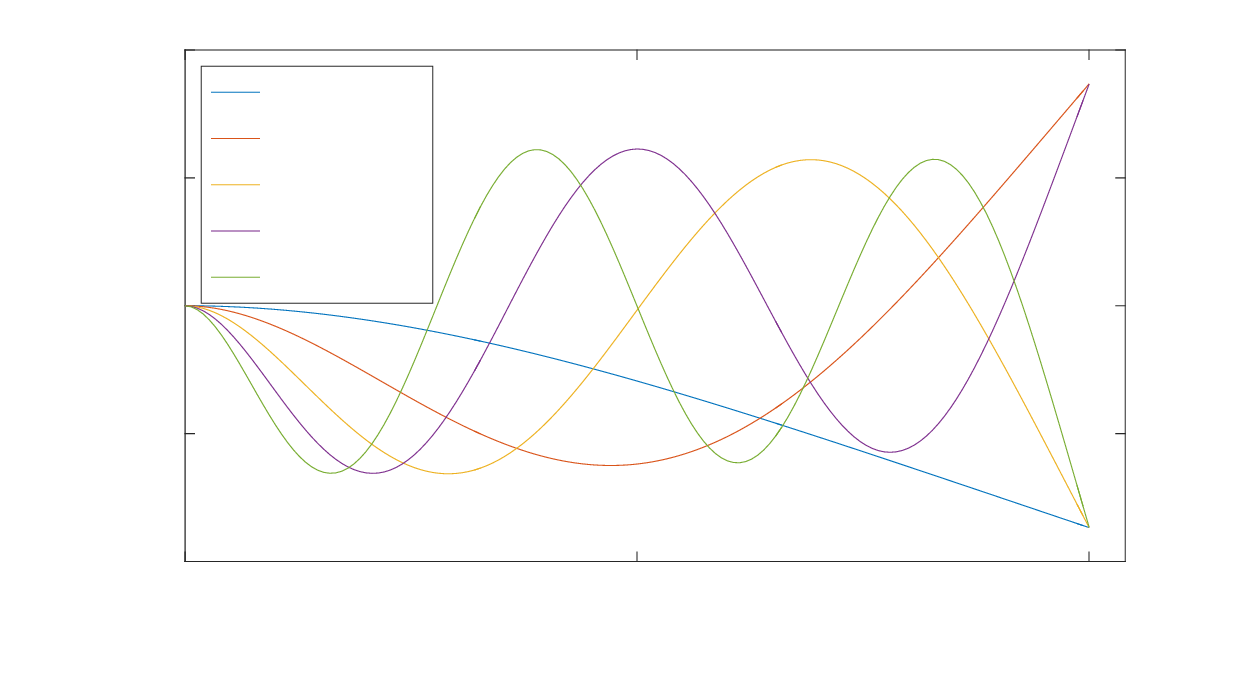
\caption{}
\end{subfigure}
\begin{subfigure}{\linewidth}
\centering
\def\svgwidth{0.825\textwidth}
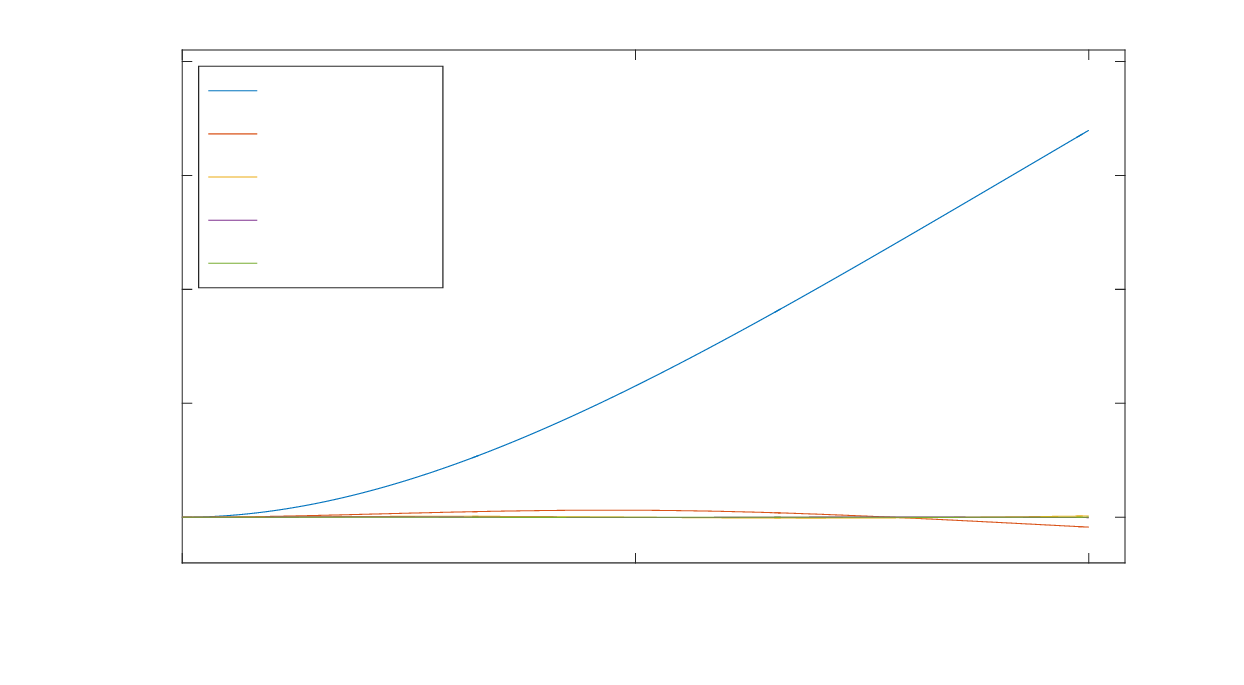
\caption{}
\end{subfigure}
\begin{subfigure}{\linewidth}
\centering
\def\svgwidth{0.6\textwidth}
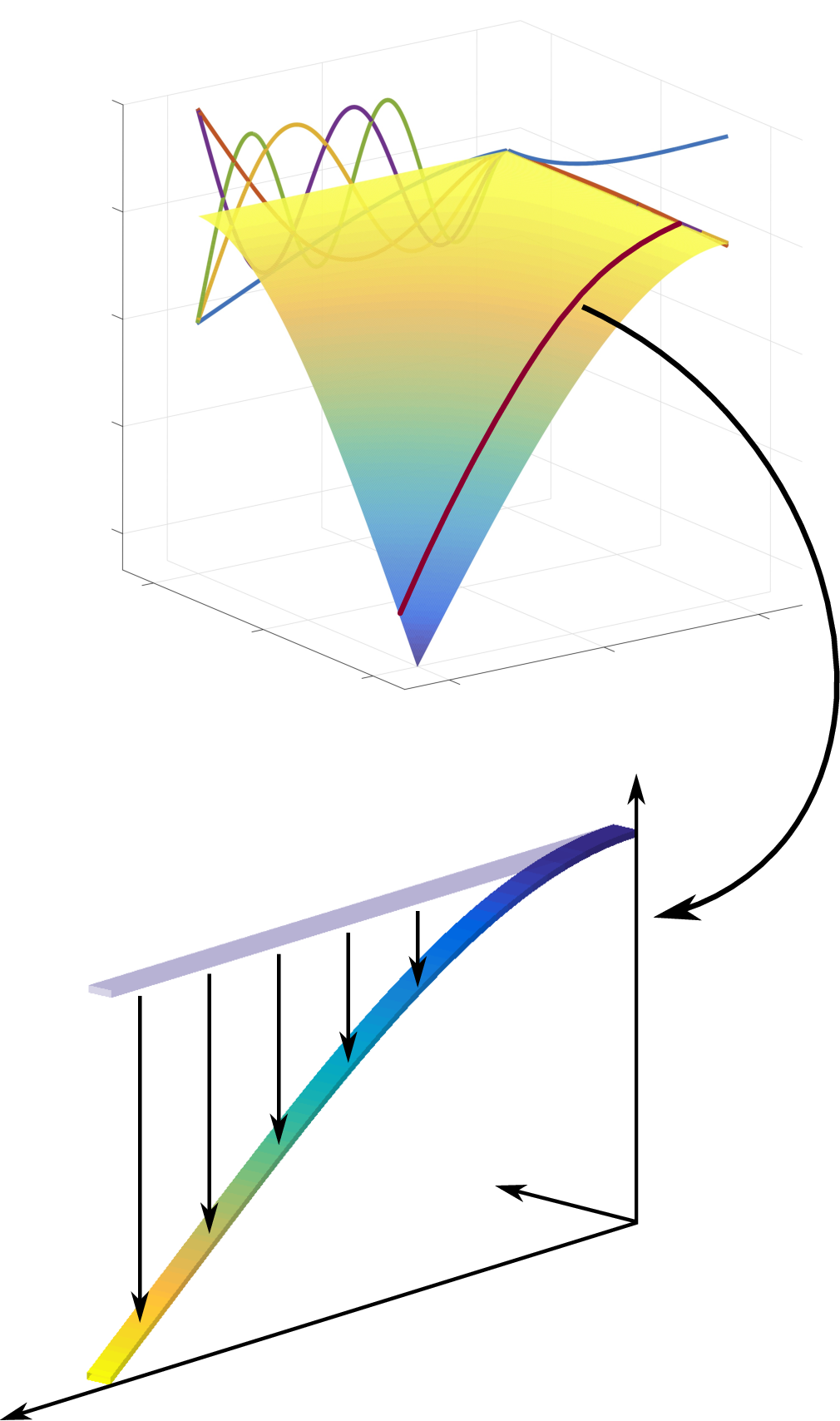
\caption{}
\end{subfigure}
}
\caption{Separated solution of the vertical displacement. (a) First five modes depending on space. (b) First five modes depending on the position of the load. (c) Built solution using the sum of the product of modes. Note how the result of the PGD approximation provides with a sort of response surface, without the need of any parameter space sampling strategy. Finally, a slice of the surface with the prescribed load position $s_0$ builds the solution.}\label{FigExampleHTE3}
\end{figure}

Therefore, the test function is
\begin{align*}
\bs U^{*}(\bs{X},s) = \bs R^{*}(\bs{X}) \circ \bs S(s) + \bs R(\bs{X}) \circ  \bs S^{*}(s).
\end{align*}
The load $F$ has been designed as a unitary force acting along the vertical axis $Y$,
\begin{align*}
\bs{t} = F \hspace{2pt} \delta (X - s) \bs{e}_{Y},\;\; X\in \Gamma_t,
\end{align*}
where $\delta$ is the Dirac delta function. Function $\bs t$ needs also to be expressed in separate form to comply with the PGD formalism, i.e.,
\begin{align*}
t_{Y} \approx \sum_{j=1}^{m} h_{j}(\bs{X}) \hspace{2pt} k_{j}(s); \hspace{15pt} t_{X} = 0,
\end{align*}
where $m$ is the number of sums to approximate function $t_{Y}$ and $h_{j}(\bs{X})$, $k_{j}(s)$ are the functions in space and load position, respectively. To solve the non-linear product of functions $\bs R(\bs{X})$ and $\bs S(s)$, the algorithm uses an alternating direction scheme based on the fixed point method at every enrichment iteration. For a detailed description of the problem and a Matlab implementation, the interested reader can consult \cite{Cueto:2016aa}.

The first modes of $\bs F(\bs{X})$ (only its vertical component, since it is a vector-valued displacement field) and $\bs G(s)$ are shown in Fig.~\ref{FigExampleHTE3}.(a) and \ref{FigExampleHTE3}.(b), respectively. Finally, the sum of the products of pairs of modes builds the approximate solution, as it is shown in Fig.~\ref{FigExampleHTE3}.(c).

It is important to highlight the fact that the accuracy of the proposed reduced-order method can be controlled by the user. The number of terms in Eq. (\ref{aa}), ${\tt n}_{\text{MOR}}$ controls this level of accuracy, with the help of a suitable error estimator, as the ones proposed in \cite{Alfaro:2015aa} \cite{Moitinho} \cite{PGDerror}, for instance.

\section{A deformable implementation of ORB-SLAM2}
\label{Sec:DefORBSLAM}
We propose in this section a deformable implementation of the ORB-SLAM2 system \cite{Mur-Artal:2017aa}. ORB-SLAM2 is a sparse and feature-based SLAM system with outstanding results of accuracy \cite{Mur-Artal:2015aa}. It is used to build a rigid scene and locate the camera in any frame. In our work, after a few seconds of video capturing the static scene, we get the 3D point cloud and apply an automatic and on-line registration against a CAD model of the object, based in RANSAC methods and the ICP (Iterative Closest Point) algorithm \cite{Besl:1992aa}.

The registration step allows us to know which points are deformable and, subsequently, to estimate the displacements of the 3D SLAM points ($\bs{U}_{\text{SLAM}}(\bs{X},\bs{\mu)}$) using the precomputed data from the CAD object
\begin{equation}
\bs{U}_{\text{SLAM}}(\bs{X},\bs{\mu}) =  f \Big( \bs{U}_{\text{Neighs,CAD}}(\bs{X},\bs{\mu}) \Big).
\end{equation}

\begin{figure}[!h]
\centering
\scriptsize{
\def\svgwidth{0.45\textwidth}
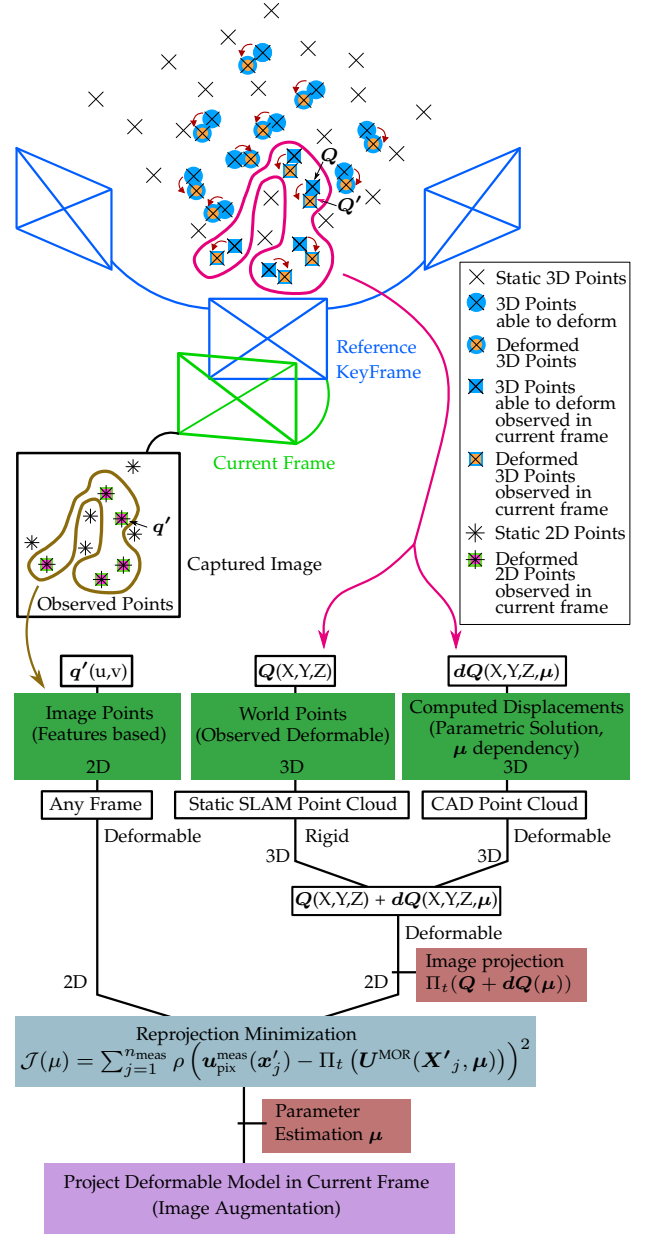
}
\caption{Deformable object tracking and augmentation procedure based in ORB features matching.}
\label{FigORBMatching}
\end{figure}

To do that, a mapping function between our undeformed 3D model and the 2D image features need to be established. It means we can afterwards deform the real object, extract image features in any frame and estimate the parameter values $\bs{\mu}$ {  with respect to the precomputed set of solutions.} The goal is to obtain the displacements $\bs{U}_{\text{SLAM}}(\bs{X},\bs{\mu})$ to add to the static set of deformable 3D points initially estimated using the SLAM technique. This solution must maintain the separate multiparametric structure to obtain a specific solution for each possible value of our set of parameters. Therefore, the displacements of the SLAM points are expressed as a function of the closest registered CAD points. The procedure can be seen in the general overview schema (Fig.~\ref{FigORBMatching}).

{ In the visual scanned point cloud there are 3D points that remain static (used to fix camera position and orientation) and 3D  points that are assigned to the deformable part of the domain in the current observation. Let us focus on the points belonging to the deformable part, $\bs{Q}$, with coordinates $(X,Y,Z)$. }The position in its deformed configuration is $\bs{Q'}$, {  where} $\bs{U} = (\bs{X'} - \bs{X}) = (X', Y', Z') - (X, Y, Z) $, but this displacement $\bs{U}$ is actually estimated from the deformations of the CAD object, which depend on the multiparametric solution projected in separate variables. The point in its deformed configuration ($\bs{Q'}$) is observed from the image as $\bs{q'}$ with pixel coordinates $(u', v')$, which must match the projection $\bs{q}'_{e} = \Pi_{t}(\bs{Q} + \bs{U}(\bs{\mu}))$ for the optimal set of parameters $\bs{\mu}$. The reprojection error in the image is therefore defined as $d(\bs{q'},\bs{q'_{e}})$  and the optimal parameter values of our parametric solution that minimize this error for the whole set of observed points are obtained from the next functional
\begin{align}
 \mathcal{J}(\bs{\mu}) = \sum_{j=1}^{\tt n_{\text{meas}}} \rho \Big( \bs u^{\text{meas}}_{\text{pix}} (\bs x'_j) -  \Pi_{t} \left(\bs U ^{{\text{MOR}}} (\bs{X}'_j, \bs{\mu})\right) \Big)^2
 \label{eq:C4:functional}
\end{align}
{  where $\tt n_{\text{meas}}$ are} the number of measurements, $\bs u^{\text{meas}}_{\text{pix}}$ the pixel coordinates measured in the image, $\bs U^{\text{MOR}}(\bs{X}'_{j}, \bs{\mu})$ the 3D displacement predicted by the reduced-order parametric solution, $\Pi_{t}$ the projection to the image plane at any  instant $t$ (frame), and $\rho$ the Huber robust function. It is important to highlight that $\bs U ^{{\text{MOR}}} (\bs{X}'_j, \bs{\mu})$ represents the whole solution of our hyperelastic problem in question by storing the position, deformations and stresses of each point, with parametric dependence and represented in the reduced space. However, to simplify the process, $\bs{U}$ in the functional defined in Eq.~(\ref{eq:C4:functional}) only represents the deformed position of each point. Using Eq.~(\ref{eq:C4:functional}) we can estimate the value of the $\bs{\mu}$ parameters that minimize the reprojection error
\begin{align*}
\hat{\bs{\mu}} = \argmin_{\bs{\mu}} \mathcal J(\bs{\mu}).
\end{align*}
This minimization can be advantageously accomplished by employing the Levenberg-Marquardt algorithm \cite{Levenberg:1944aa}\cite{Marquardt:1963aa}, which takes the form
\begin{multline*}
[\bs{J}^{T} \bs{J} + \lambda \hspace{2pt} \text{diag}(\bs{J}^{T} \bs{J})] \hspace{2pt} \delta  \\ 
= \bs{J}^{T} \left [\bs U^{\text{meas}}_{\text{pix}} (\bs x'_j) -  \Pi_{t} \left(\bs U ^{{\text{MOR}}} (\bs{X}'_j, \bs{\mu})\right) \right ],
\end{multline*}
where $\delta$ is the perturbation parameter that moves the parameters in the direction of steepest descent and $\lambda$ an algorithmic parameter that adaptively varies the behavior between the Gradient Descent Method and the Gauss-Newton Method. Finally, $\bs{J}$ is the Jacobian matrix, defined as
\begin{equation}
\bs{J} = \frac{\partial \bs{U}(\bs{x}, \bs{\mu})}{\partial \bs{\mu}}.
\label{Eq:MatJacob}
\end{equation}
Here lies precisely the key point that makes the separate representation of the solution so powerful in the parameter optimization process. Recall the general form of PGD, Eq.~(\ref{sep2}), where we can notice that due to its separate form, a separate differentiation can also be performed. We can precompute the derivative function of each parameter, Eq.(~\ref{Eq:MatJacob}), offline, since the finite element approximation computes the derivative as a multiplication of matrices. These derivative functions are usually known as \emph{sensitivities}  of the solution \cite{Chinesta:2013ab} with respect to each parameter $\mu_{k}$.  We can interpret them as the local variation of the function $\bs U(\bs{X}, \bs{\mu})$ with respect to the changes in parameter ${\mu}_k$
\begin{align*}
\bs{J}_{k} = \sum_{i=1}^{N} \bs F_i^1(\bs{X})\circ \ldots \circ \frac{\partial \bs{F}_{i}^{k}( {\mu}_{k})}{{\mu}_{k}} \circ  \hspace{1pt} {\ldots} \circ  \bs{F}_{i}^{\tt n_{\text{param}}}({\mu}_{\tt n_{\text{param}}}).
\end{align*}

It allows us to avoid the exploration of all of the parametric space in search of the derivatives, and directly apply the differentiation on the separated variable vectors, involving a great reduction in the computational task.
\begin{figure}[h]
\centering
\def\svgwidth{0.44\textwidth}
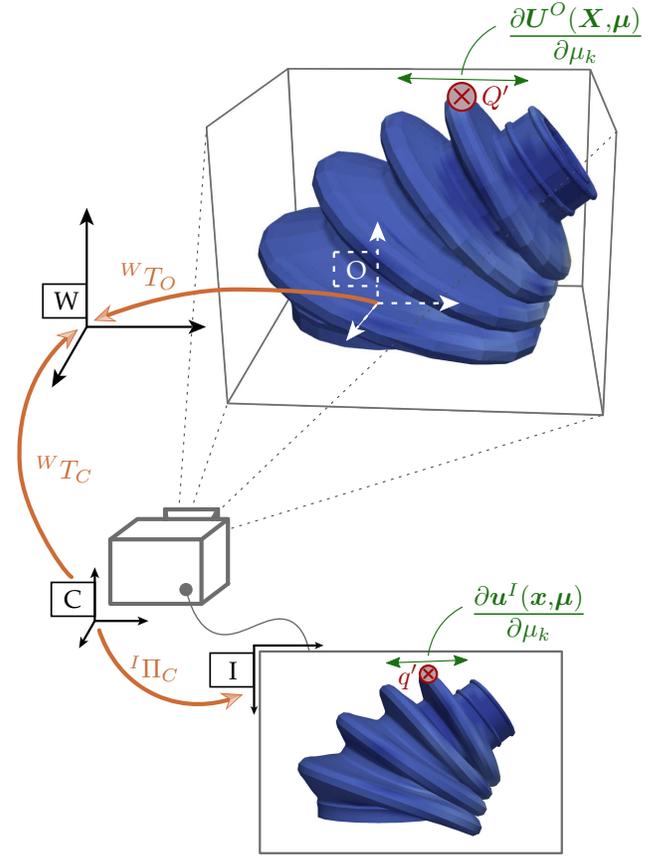
\caption[Sensitivities transformations and projection.]{Transformations and projection of the precomputed sensitivities from the original system of coordinates to the image reference.}
\label{fig:C4:ProjectSensitivities}
\end{figure}
In sum, we need the partial components of the derivative terms of the multiparametric solution projected on the image, since we are minimizing the reprojection error (2D), but we precomputed the sensitivities in the original 3D space, so we have to project these sensitivities onto the image plane. A graphic example to understand this process appears in Fig.~\ref{fig:C4:ProjectSensitivities}, where
\begin{align}
\bs{J}_{k} = \frac{\partial \bs{U}^{O}(\bs{X},\bs{\mu})}{\partial \mu_{k}}
\label{Eq:3DSensitivity}
\end{align}
refers to the partial derivative term of the multiparametric 3D solution of the CAD object with respect to the parameter $\mu_{k}$, evaluated at point $\bs{Q'}$ and with the origin of coordinates defined on that object ($O$), and
\begin{align}
\bs{j}_{k} = \frac{\partial \bs{u}^{I}(\bs{x},\bs{\mu})}{\partial \mu_{k}}	
\label{Eq:ImageSensitivity}
\end{align}
represents the same partial derivative projected on the image plane and evaluated on the point $\bs{q'}$. To obtain the value of this sensitivity in the image plane, $\bs{j}_{k}$, we use the chain rule
\begin{align*}
\bs{j}_{k} = \frac{\partial \bs{u}^{I}(\bs{x},\bs{\mu})}{\partial \bs{U}^{C}(\bs{X},\bs{\mu})} \frac{\partial \bs{U}^{C}(\bs{X},\bs{\mu})}{\partial \bs{U}^{W}(\bs{X},\bs{\mu})} \frac{\partial \bs{U}^{W}(\bs{X},\bs{\mu})}{\partial \bs{U}^{O}(\bs{X},\bs{\mu})} \frac{\partial \bs{U}^{O}(\bs{X},\bs{\mu})}{\partial \mu_{k}},
\end{align*}
{  where $\bs{u}^{I}(\bs{x},\bs{\mu})$ is} the position of point $\bs{q'}$ with respect to the image plane, $\bs{U}^{C}(\bs{X},\bs{\mu})$ with respect to the camera coordinates, $\bs{U}^{W}(\bs{X},\bs{\mu})$ with respect to the global coordinate axis (\emph{World}) and $\bs{U}^{O}(\bs{X},\bs{\mu})$ with respect to the coordinate axis defined in the object. The solution derivative terms are known in the object reference, Eq.~(\ref{Eq:3DSensitivity}), and their consecutive transformations in the Euclidean space $\mathbb{R}^3$ are carried out by the matrices ${^{W}\bs{T}_{O}}$ and ${^{W}\bs{T}_{C}}$, which are assumed to be known since they are estimated by the static part of ORB-SLAM2 and the registration step. However, the projection on the image plane ${^{I}\Pi_{C}}$ does not allow such a simple treatment, since the camera conical projection used depends on the location of the 3D point with respect to the image plane. This means that the projection of the derivative terms $\mathbb{R}^3 \mapsto \mathbb{R}^2$ must take into account the position of the point where the gradients are evaluated. This projection is known as the image Jacobian, and is expressed as
\begin{align*}
{^{I}\bs{J\Pi}_{C}} = \frac{\partial \bs{u}^{I}(\bs{x},\bs{\mu})}{\partial  \bs{U}^{C}(\bs{X},\bs{\mu})} =
\begin{bmatrix}
\frac{1}{Z_{C}}f_{x} &                     0 &   -\frac{X_{C}}{Z_{C}^{2}}f_{x}\\
                  0  &  \frac{1}{Z_{C}}f_{y} &   -\frac{Y_{C}}{Z_{C}^{2}}f_{y}\\
\end{bmatrix},
\end{align*}
where $({X_{C}},{Y_{C}},{Z_{C}})$ are the 3D coordinates of point $\bs{Q}$ with respect to the camera axes. Finally, the projection of the multiparametric derivative terms with respect to parameter $\mu_{k}$ on the image plane is expressed as
\begin{align}
\frac{\partial \bs{u}^{I}(\bs{x},\bs{\mu})}{\partial\mu_{k}} = \hspace{2pt} {^{I}\bs{J\Pi}_{C}} \cdot  \hspace{2pt} {^{W}\bs{T}_{C}}^{-1} \cdot \hspace{2pt} {^{W}\bs{T}_{O}} \cdot \frac{\partial \bs{U}^{O}(\bs{X},\bs{\mu})}{\partial \mu_{k}},
\label{Eq:C4:ProjPartialDeriv}
\end{align}
where we remember that the sensitivities defined with respect to the object axes are interpolated according to the neighbors in the separate variables precomputed solution. Regarding to the technical aspects, we use the feature search method developed in the ORB-SLAM2 code to track the deformations. Time requirements for the minimization algorithm are much smaller than the computation time of the SLAM tasks, so a simple parallelization of the whole code in a few threads allows us to work in a frequency of 30~fps using the CPU. We stop the \emph{Mapping} tasks and work in \emph{Localization} mode in ORB-SLAM2 once deformations are appearing.

\section{Experiments}
\label{Sec:Experims}
We provide two examples to support our method. Both have been tested with different video sequences working in real time.

\subsection{Rubber boot seal}
\label{Eq:ExamplesRubberPiece}

This example shows the behavior of a rubber boot seal, the part that protects the gearshift of a vehicle. The object is deformed forced by the rotation of the gearshift in one direction, from $-52^\circ$ to $52^\circ$, but the simulation has been only applied within the parametric range $\theta \in [0^\circ,52^\circ]$, taking advantage of revolution symmetry.

The material behavior applied to the rubber is a non-linear law, called Neo-Hookean model, with the strain energy density
\begin{equation*}
W = \frac{\mu}{2}(I_{1} - 3) - \mu \ln(J) + \frac{\lambda}{2} (\ln (J))^2,
\end{equation*}
where $I_{1}$ is the first invariant of the right Cauchy-Green deformation tensor
\begin{equation*}
I_{1} = \lambda_{1}^{2} + \lambda_{2}^{2} + \lambda_{3}^{2},
\end{equation*}
and where $\lambda_{i}$ are the eigenvalues. The mesh of the object has 25795 nodes and 18403 linear hexahedral elements.

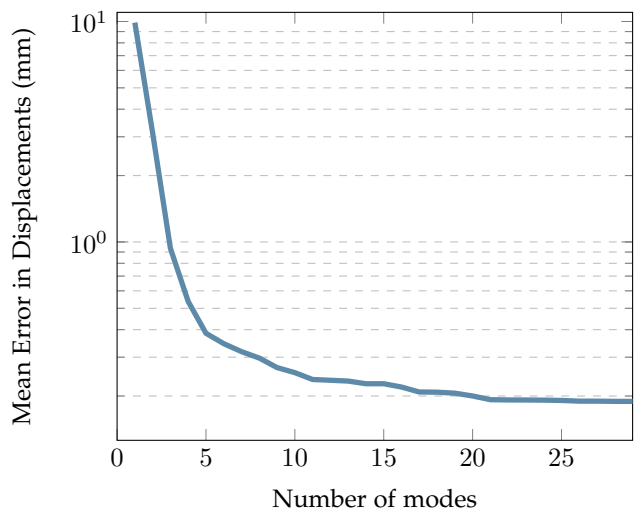
\begin{figure}[!h]
\centering
	\begin{tikzpicture}
		\begin{semilogyaxis}[
	  		scale = 1,			    
		    xlabel={Number of modes},
		    ylabel={Mean Error in Displacements (mm)},
		    xmin=0, xmax=29,
		    ymin=0, ymax=11,
		    xtick={0,5,10,15,20,25},
		    ytick={0.1,1,10},
		    yminorgrids=true,
		    grid style=dashed,
		] 

				\addplot[
			    color=airforceblue,
			    line width=2pt,
			    ]
			    coordinates {(1,9.8840)    (2,3.1217)    (3,0.9354)    (4,0.5369)    (5,0.3853)    (6,0.3456)    (7,0.3180)    (8,0.2973)    (9,0.2692)    (10,0.2553)    (11,0.2377)    (12,0.2358)    (13,0.2338)    (14,0.2274)    (15,0.2274)    (16,0.2197)  (17,0.2086)    (18,0.2082)    (19,0.2060)    (20,0.2001)    (21,0.1925)    (22,0.1920)    (23,0.1919)    (24,0.1916)    (25,0.1910)    (26,0.1896)    (27,0.1896)    (28,0.1892)    (29,0.1891)};
		\end{semilogyaxis}
	\end{tikzpicture}
\caption{Cartesian mean 3D error in displacements due to the projection of the solution in the reduced space respect to the number of modes.}
\label{Fig:ModesMeanCartesianError}
\end{figure}

Applying a sparse implementation of PGD \cite{Ibanez2018}, we obtain the parametric solution in the projected space, depending in space $\bs{X} = (X,Y,Z)$ and one parameter, $\theta$, where the parameter optimization is a very fast process as we precompute the sensitivities. Due to the projection process of the data, a certain error in the approximation is made, as it can be seen in Fig.~\ref{Fig:ModesMeanCartesianError}. With a total number of 29 modes, we get a cartesian mean error of 0.18 mm (the normalized error is 0.0041). For this solution, the compression factor is 92.58\%, computed as
\begin{equation*}
C(\%) = \Big(1 - \frac{M_{P}}{M_{O}} \Big) \cdot 100,
\end{equation*}
where $M_{P}$ is the memory storage used by the projected solution and $M_{O}$ is the memory cost necessary to store the original data. In other words, using a reduced space to express our solution we only need a 7.42\% of the storage memory, allowing real time evaluations.

\begin{figure}[H]
\centering
  	\def\svgwidth{0.5\textwidth}
\begingroup%
  \makeatletter%
  \providecommand\color[2][]{%
    \errmessage{(Inkscape) Color is used for the text in Inkscape, but the package 'color.sty' is not loaded}%
    \renewcommand\color[2][]{}%
  }%
  \providecommand\transparent[1]{%
    \errmessage{(Inkscape) Transparency is used (non-zero) for the text in Inkscape, but the package 'transparent.sty' is not loaded}%
    \renewcommand\transparent[1]{}%
  }%
  \providecommand\rotatebox[2]{#2}%
  \newcommand*\fsize{\dimexpr\f@size pt\relax}%
  \newcommand*\lineheight[1]{\fontsize{\fsize}{#1\fsize}\selectfont}%
  \ifx\svgwidth\undefined%
    \setlength{\unitlength}{1615.84204102bp}%
    \ifx\svgscale\undefined%
      \relax%
    \else%
      \setlength{\unitlength}{\unitlength * \real{\svgscale}}%
    \fi%
  \else%
    \setlength{\unitlength}{\svgwidth}%
  \fi%
  \global\let\svgwidth\undefined%
  \global\let\svgscale\undefined%
  \makeatother%
  \begin{picture}(1,0.50381454)%
    \lineheight{1}%
    \setlength\tabcolsep{0pt}%
    \put(0,0){\includegraphics[width=\unitlength]{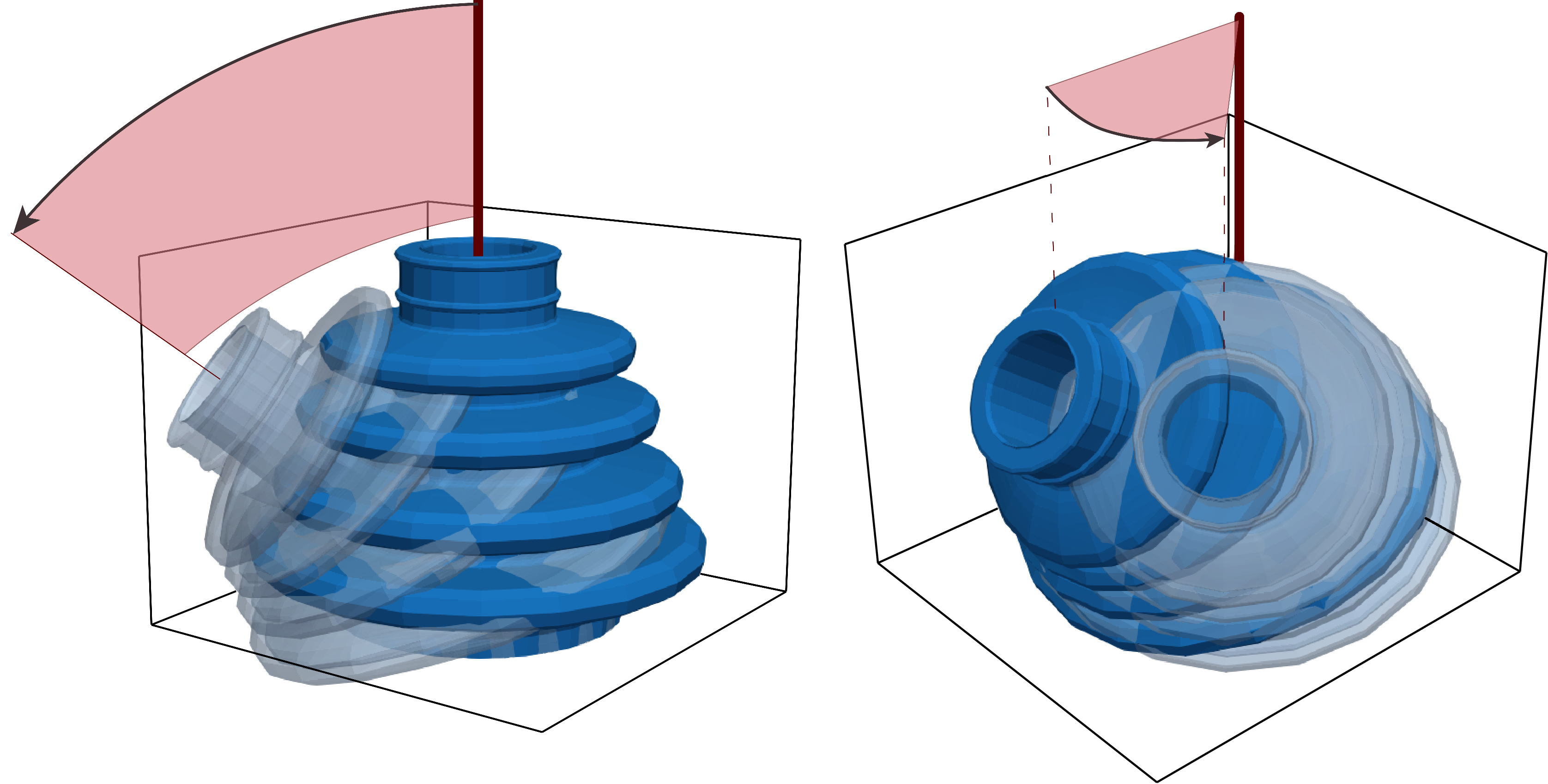}}%
    \put(0.72593063,0.38680943){\color[rgb]{0,0,0}\makebox(0,0)[lt]{\lineheight{1.25}\smash{\begin{tabular}[t]{l}\textcolor[rgb]{0.247,0.196,0.2039}{$\alpha_{Z}$}\end{tabular}}}}%
    \put(0.09335392,0.45829704){\color[rgb]{0,0,0}\makebox(0,0)[lt]{\lineheight{1.25}\smash{\begin{tabular}[t]{l}\textcolor[rgb]{0.247,0.196,0.2039}{$\theta$}\end{tabular}}}}%
  \end{picture}%
\endgroup%

\caption{Parameters to optimize in the video sequence.}
\label{FigParamsRubber}
\end{figure}

\begin{figure*}[!ht]
\centering
\includegraphics[width=\textwidth]{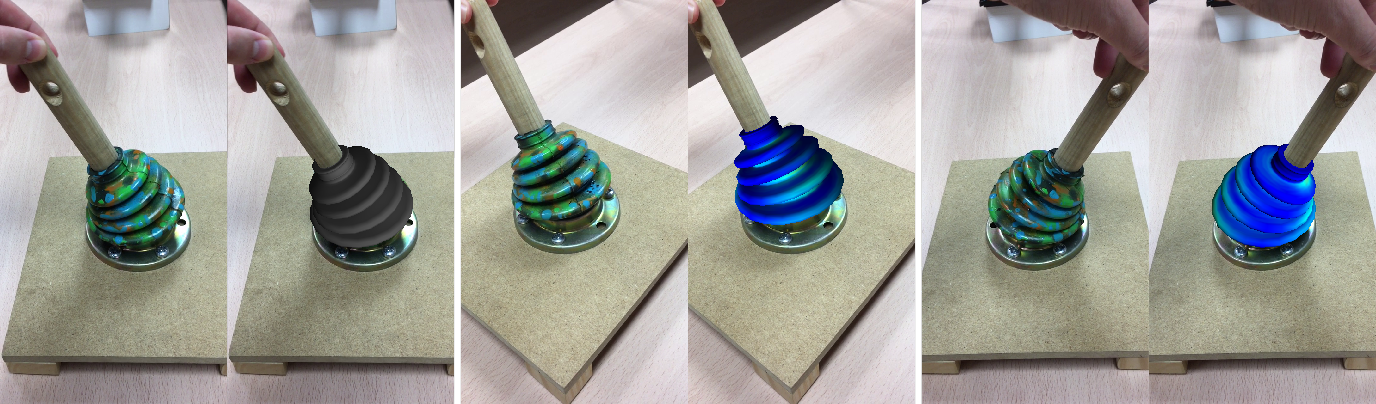}
\caption{Some images showing the original and the augmented frames. In the central and right images we can see the stress field that the object is suffering when the deformation appears.}
\label{FigBootSealAR}
\end{figure*}

Since we assume that the geometry of the deformable model is known, but not its external appearance (texture) or its location in the scene, we apply a registration process between the static point cloud and the CAD object. { The object to be deformed has geometry of revolution, so we must actually optimize the value of two parameters $\bs{\mu} = (\theta, \alpha_{Z})$, where parameter $\theta$ measures the deformation and $\alpha_{Z}$ the angle of orientation of the axis of rotation, a geometric parameter that we cannot estimate in the registration process (see Fig.~\ref{FigParamsRubber}).} This is why, when the deformation occurs, we have to estimate the value of $\theta$ in each frame. 
\begin{figure}[!hb]
\centerline{\includegraphics[width=0.85\columnwidth]{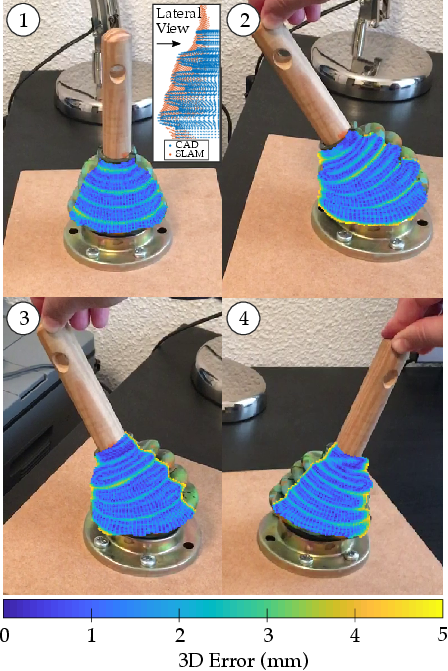}}
\caption{Error of the monocular estimation, measured assuming the ground-truth is coming from a stereo camera. Due to the high complexity of the geometry, the stereo system can not be more accurate than the sum of model and monocular camera in some areas, producing a constant error that can be shown in the undeformed frame \raisebox{.5pt}{\textcircled{\raisebox{-.9pt} {1}}}.}
\label{FigErrorTest}
\end{figure}
To favor the minimization process, the value of $\alpha_{Z} $ is minimized throughout the entire sequence, so after a few frames observing the deformed object, the system itself is able to estimate with great degree of certainty the value of $\alpha_{Z}$, which involves also a better accuracy in the estimation of parameter $\theta$.

\begin{figure}[!h]
\centering
\def\svgwidth{0.4\textwidth}
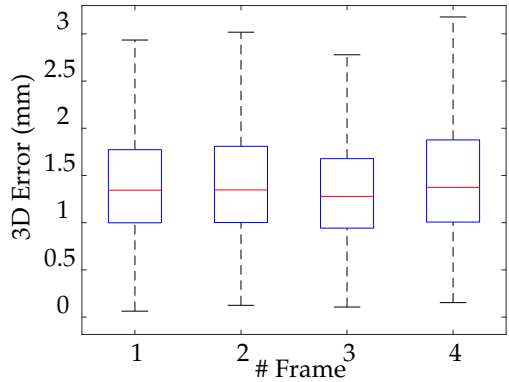
\caption{Boxplot of the error distribution of all points of the external geometry, acquired in the frames of Fig.~\ref{FigErrorTest}.}
\label{FigErrorBoxPlot}
\end{figure}

Fig.~\ref{FigBootSealAR} shows some frames extracted from a video sequence with the original object, and the virtual object with the estimated displacements. The central and right frames show the stress field in colors, result of solving the whole mechanical problem. Finally, in Figs.~\ref{FigErrorTest} and \ref{FigErrorBoxPlot} the 3D error is estimated. These error values have been calculated as the euclidean distances between the surface of the deformed virtual object and the object scanned with an RGB-D camera (Intel RealSense D415). { It is important to note that using an accurate model of a complex geometry like the one in this example gives better results than the scanning of an RGB-D system, mainly due to the precision of the camera itself. For example, it can be seen in image 1 of Fig.~\ref{FigErrorTest} that the RGB-D camera is not able to accurately estimate the folds of the real object. In any case, the median error using a monocular camera is 1.33 mm (Fig.~\ref{FigErrorBoxPlot}) and the mean error is 1.49 mm. These very accurate values are the result of using a detailed model of the geometry, an adequate description of the material law, and a measurement system as accurate as the non-deformable implementation of ORB-SLAM2.}

\begin{figure*}[!ht]
\centering
\includegraphics[width=\textwidth]{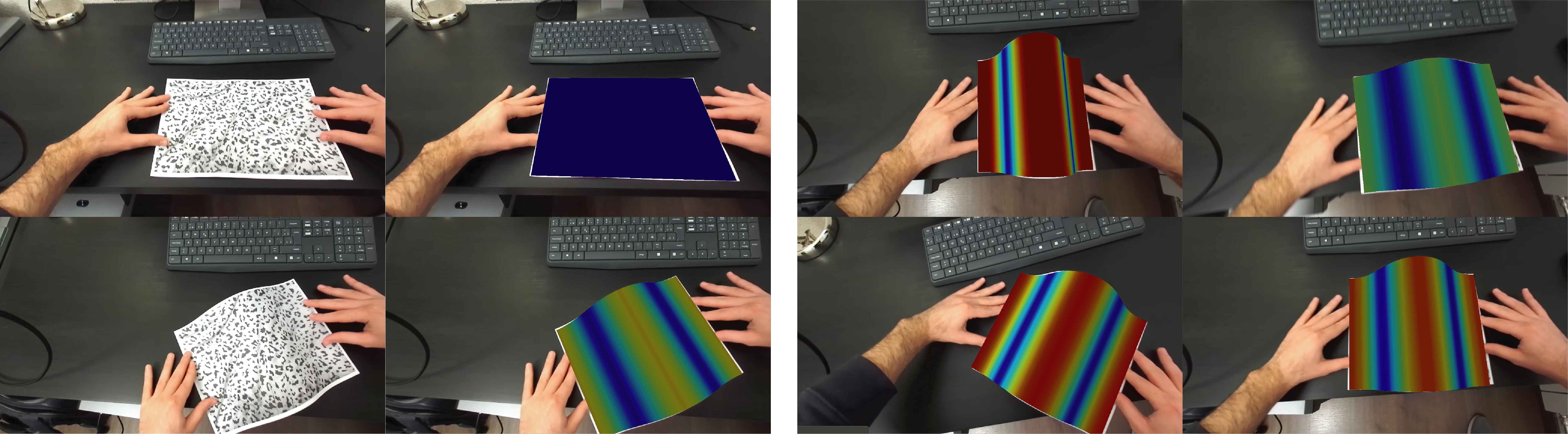}
\caption{{ Some images showing the original and the augmented frames in the second example. Colors in images represent the real-time stress field that the object is suffering when the deformation appears.}}
\label{FigPaperSheet}
\end{figure*}

{ 
\subsection{Paper sheet buckling}
\label{Eq:ExamplesPaperSheet}

The second example simulates the buckling phenomenon of a paper sheet. From a numerical simulation point of view, buckling is a complex problem that is usually solved using linear approximations to obtain the critical loads (buckling modes) that produce the beginning of buckling. For this, an eigenvalue analysis is used, looking for the loads for which the model stiffness matrix becomes singular \cite{Novoselac:2012aa}. However, if a complete analysis of the buckling problem is desired, in order to know in detail the values of the stresses and deformations of the object in question, it is necessary to apply more complex methods such as the Riks method, introducing small imperfections on the geometry \cite{Novoselac:2012aa}.

Again, the hypothesis of the use of a reduced model in this example is based on the fact that the number of degrees of freedom of the problem is not high. In other words, although it may appear that the object is arbitrarily deformed, there is a great restriction due to the geometry of the paper, so the number of degrees of freedom of the problem is very small, actually living in a reduced dimension manifold.

The paper sheet material has been considered anisotropic (reinforced with oriented fibers) to emulate the real behavior of the object \cite{Schulgasser:1981aa, Yokoyama:2007aa}. The model mesh has been discretized in 1333 nodes, 31 in the vertical direction and 43 in the horizontal direction. To model the buckling problem we use one single parameter ($\mu_{B}$) discretized at 801 buckling positions. Applying again the sparse implementation of PGD \cite{Ibanez2018} it is possible to obtain a projected solution $\bs{u} = (X,Y,\mu_{B})$, reaching a compression factor of 98.36\%.

The sequence was recorded with an RGB-D camera (StereoLabs Zed Mini) in order to estimate the error made by our monocular method, where Fig.~\ref{FigErrorPaperSheet} plots the mean error in millimeters between the model and the real object for each frame. Fig.~\ref{FigPaperSheet} shows some frames extracted from the sequence, where colors represent the Von Mises stress field. Note that this error, that may seem big, is comparable to the best of the state-of-the-art, with one important difference: we are solving the physics of the problem, and therefore are subject to errors in the calibration of material parameters, for instance \cite{Lamarca2019}. In addition, deformations in the considered sequence are much bigger than those in \cite{Parashar:2021aa}, with a maximum vertical displacement value (also corresponding to the maximum error peak) of 16.9 cm. Relative error is thus on the order of 10\% at this maximal value.

\begin{figure}[!h]
  \centerline{\includegraphics[width=0.45\textwidth]{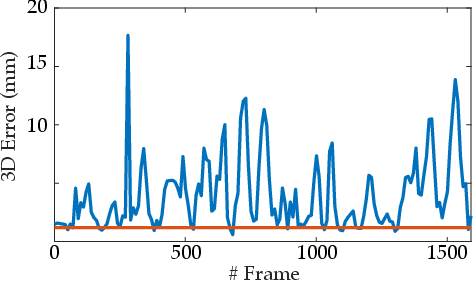}}
\caption{{  Mean 3D error of our monocular method compared to ground truth (RGB-D camera) for the whole sequence. Orange line shows the camera mean error (base error) according to our experiments.}}
\label{FigErrorPaperSheet}
\end{figure} 

The fact that both deformations and displacements (as a rigid solid) of the object are allowed makes it also necessary to estimate the deformation parameters and the pose of the object in each frame. This means that the robustness of ORB-SLAM2 in its rigid part cannot be exploited and the errors are slightly higher than those of the previous example, but our method is more general and robust to displacements. The 3D mean error of the whole sequence is 3.98 mm, measured respect to the ground truth data coming from the RGB-D camera and taking into account there is a relevant camera error of 1.2 mm (orange line in Fig.~\ref{FigErrorPaperSheet}). This error is one order of magnitude less than the one reported in state of the art references, such as \cite{Lamarca2019}, for instance, that reported errors up to 30--35 mm in the displacement prediction.

}

\section{Conclusions}

We propose in this paper the application of computational mechanics as a tool for estimating the deformations suffered by the objects perceived by a standard camera. This allows us to augment the video sequence with information regarding the displacements, forces and stresses suffered by the objects. To do this, we rely on the use of models to obtain a parametric solution (using finite element methods) that we use in the on-line phase. This parametric solution is stored in a compressed way to optimize both the cost of storage and its evaluation in real time.

One of the great differences of this work with respect to the bibliography is that we do not have to impose any type of spatio-temporal restriction, nor apply energy conservation laws implemented ad-hoc, as we are solving the real physics of the deformable solids. This also implies that the result is defined for the whole volume of the modeled objects, not only for the external surface, allowing us to capture the real behavior of the objects although partial occlusions appear. It also provides information about how the interior of the objects is changing, even though we are not able to capture them with the camera. It assums a clear advantage for biomedical applications where it is necessary to know, for example, the location of the blood vessels or cancerous bodies inside the organs. Also different material properties can be applied in any point of the solid, giving different results depending on the stiffness, but assuming more expensive previous work.

As it is shown in the experiments section, our method has a great computational efficiency, is CPU-based and quite robust against noisy observations. It is not necessary to precompute sequences to extract trajectories or learning shapes, as we obtain the information directly from the models. But the solution must be obtained for a parametric set of loads, states and deformations enough to be able to work in any sequence.

Finally, this work can use any type of model order reduction method, and applied to any type of camera device, such as monocular, stereo or RGB-D systems.

\ifCLASSOPTIONcompsoc
  \section*{Acknowledgments}
\else
  \section*{Acknowledgment}
\fi

We are grateful to our colleague Prof. J. M. Mart\'{i}nez Montiel, whose help with some technical aspects is much appreciated.

This work has been partially funded by the ESI Group through the ESI Chair at ENSAM Arts et Metiers Institute of Technology, and the project 2019-0060 ''Simulated Reality" at the University of Zaragoza. We also thank the support of the Spanish Ministry of Economy and Competitiveness, grant CICYT-DPI2017-85139-C2-1-R and the Regional Government of Aragon and the European Social Fund.

\bibliographystyle{IEEEtran}
\bibliography{myBibliographyAllAsJournals}

\begin{IEEEbiography}[{\includegraphics[width=1in,height=1.25in,clip,keepaspectratio]{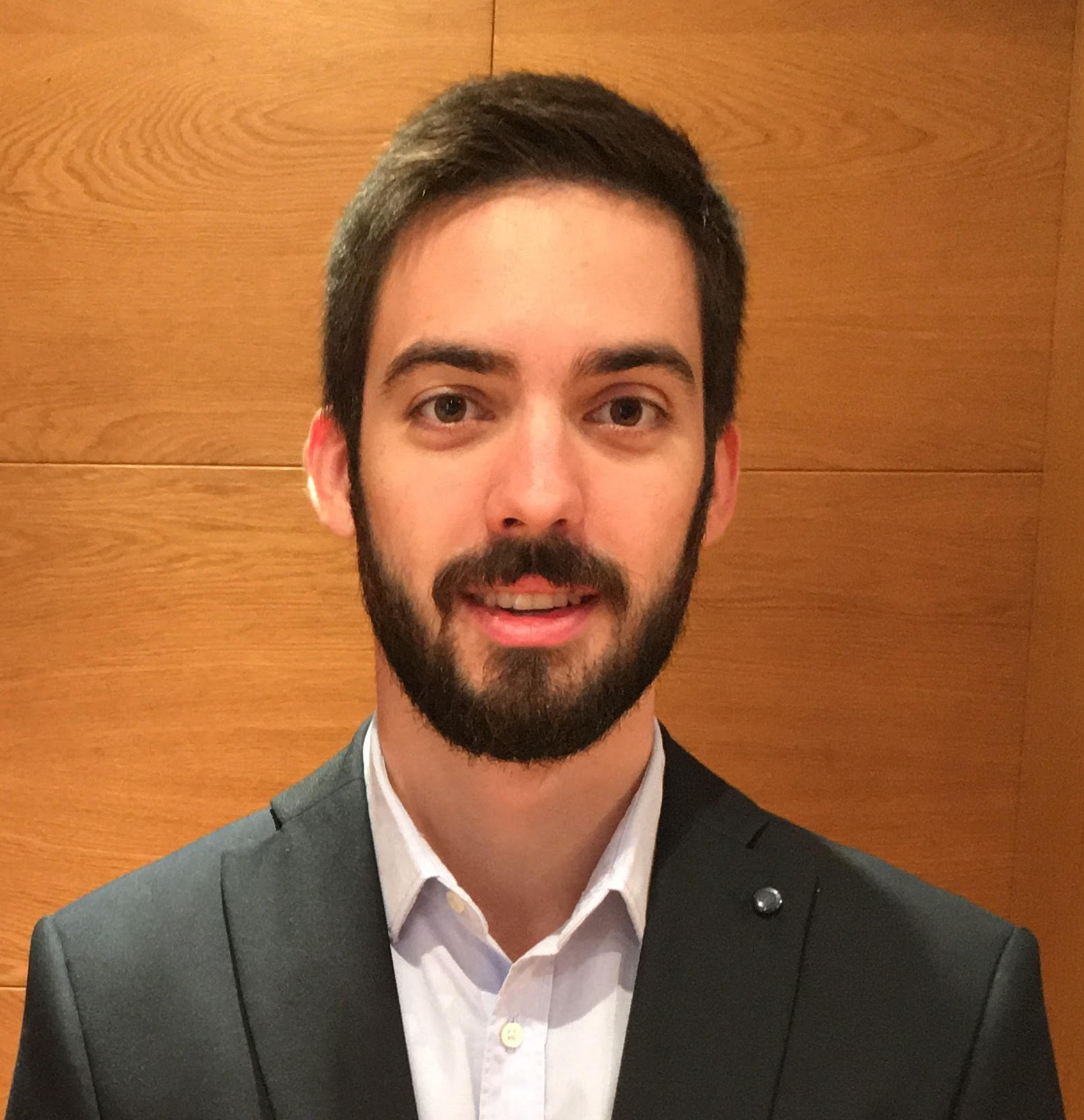}}]{Alberto Bad\'{i}as} is currently a postdoctoral researcher at the University of Zaragoza, Spain. He received the B.S. degree in mechanical engineering in 2011, the M.S. degree in industrial engineering (industrial automation and robotics) in 2014, the M.S. degree in biomedical engineering in 2016 and the Ph.D. degree in 2020, all from the University of Zaragoza. He has been working in the area of computer vision and robotics developing 3D reconstructions and new image processing tools, and is working currently in model order reduction methods in applied mechanics and bioengineering.
\end{IEEEbiography}

\begin{IEEEbiography}[{\includegraphics[width=1in,height=1.25in,clip,keepaspectratio]{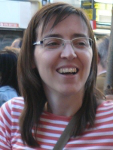}}]{Iciar Alfaro} is an associate professor of continuum mechanics at the university of Zaragoza. Her research has been devoted to the development of innovative numerical methods for complex problems in continuum mechanics. She developed natural element formulations for forming processes, and, more recently, she works in the development of advanced techniques for augmented reality with strong physical meaning.
\end{IEEEbiography}

\begin{IEEEbiography}[{\includegraphics[width=1in,height=1.25in,clip,keepaspectratio]{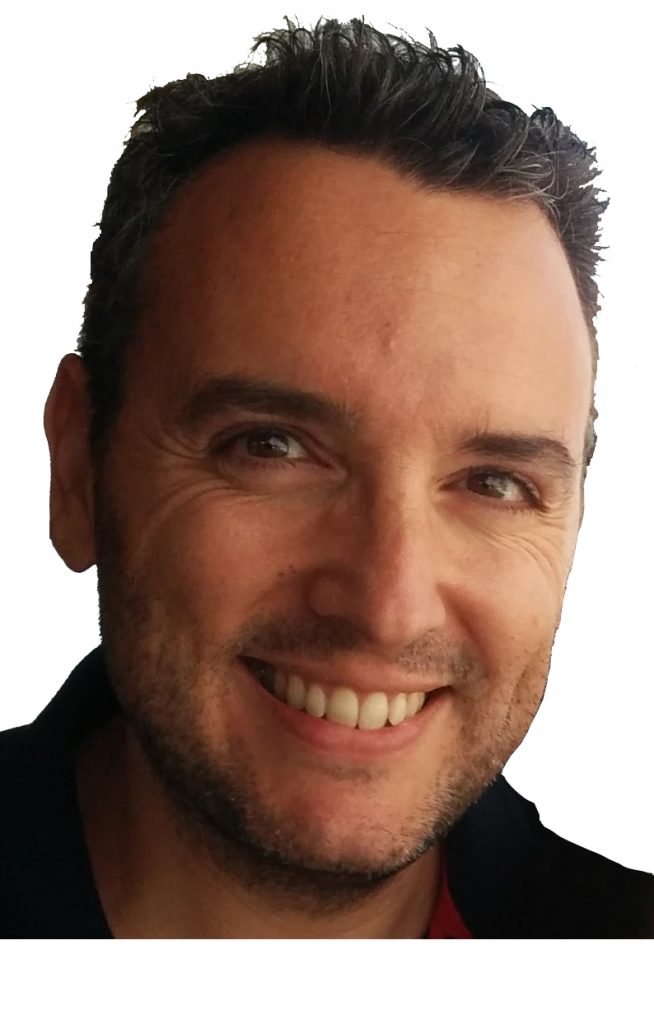}}]{David Gonz\'alez}
is currently a professor at the university of Zaragoza. He obtained his Ph.D. degree in applied mathematics in the same university in 2004, with a thesis on the development of meshless methods for the Lagrangian simulation of free-surface flows. Since then, he has been working on the development of advanced model order reduction techniques, with an eye towards their application in real-time simulation, computational surgery and augmented reality.
\end{IEEEbiography}

\begin{IEEEbiography}[{\includegraphics[width=1in,height=1.25in,clip,keepaspectratio]{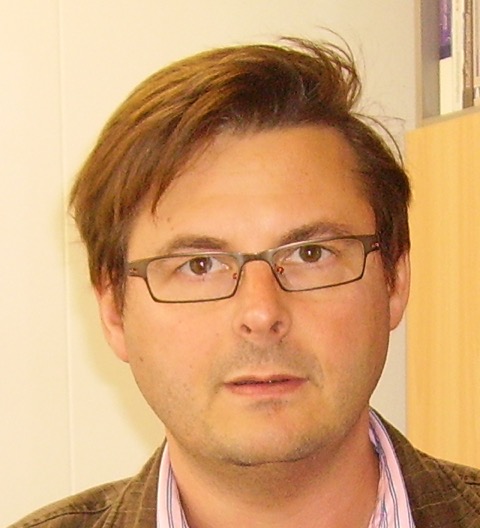}}]{Francisco Chinesta}
Francisco Chinesta is currently a professor of Computational Mechanics at the PIMM lab of ENSAM ParisTech, AIRBUS chair professor from 2008 to 2012 and ESI Chair professor since 2013. He is associate member of the University of Wales Institute of Non-Newtonian Fluid Mechanics, fellow of the ``Institut Universitaire de France'' (IUF) and fellow of the Spanish Royal Academy of Engineering. His main research contribution concerns the proposal and development of efficient model reduction strategies based on the Proper Generalized Decomposition and more particularly the development of separated representations based strategies. He is author of over 250 papers in peer reviewed journals and received many international scientific prizes and awards.
\end{IEEEbiography}


\begin{IEEEbiography}[{\includegraphics[width=1in,height=1.25in,clip,keepaspectratio]{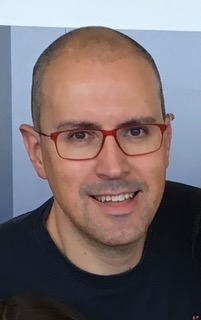}}]{El\'ias Cueto}
is professor of continuum mechanics at the university of Zaragoza. His research is devoted to the development of advanced numerical strategies for complex phenomena. In particular, in the last years he has worked in model order reduction techniques and real-time simulation for computational surgery and augmented reality applications. His work has been recognized with the J. C. Simo award of the Spanish Society of Numerical Methods in Engineering, the European Scientific Association of Material Forming (ESAFORM) Scientific Prize, and the O.C. Zienkiewicz prize of the European Community on Computational Methods in Applied Sciences, ECCOMAS, among others. He is fellow of the EAMBES society. 
\end{IEEEbiography}

\end{document}